\documentclass[times,twocolumn,final,authoryear]{elsarticle}

\usepackage[utf8]{inputenc}

\usepackage{ycviu}
\usepackage{framed,multirow}

\usepackage{amssymb}
\usepackage{latexsym}

\usepackage{amsmath}

\usepackage{url}
\usepackage{xcolor}
\definecolor{newcolor}{rgb}{.8,.349,.1}

\usepackage[capitalise]{cleveref}

\usepackage{graphicx}
\usepackage{subfigure}

\journal{Computer Vision and Image Understanding}

\begin{document}

\thispagestyle{empty}

\ifpreprint
  \setcounter{page}{1}
\else
  \setcounter{page}{1}
\fi

\begin{frontmatter}

\title{STURE: Spatial-Temporal Mutual {Representation} Learning for Robust Data Association \\in Online Multi-Object Tracking}

\author[1]{Haidong {Wang}} 
\author[1]{Zhiyong {Li}\corref{cor1}}
\cortext[cor1]{Corresponding author: 
	Tel.: +86-13607436411;}
\ead{zhiyong.li@hnu.edu.cn}
\author[1]{Yaping {Li}}
\author[1]{Ke {Nai}}
\author[2]{Ming {Wen}}

\address[1]{College of Computer Science and Electronic Engineering of Hunan University, and Key Laboratory for Embedded and Network Computing of Hunan Province, Hunan Province, Changsha and 410082, China}
\address[2]{State Grid Hunan Electric Power Company Limited Economical Technical Research Institute, Hunan Key Laboratory of Energy Internet Supply-demand and Operation, Hunan Province, Changsha and 410000, China}

\received{1 May 2013}
\finalform{10 May 2013}
\accepted{13 May 2013}
\availableonline{15 May 2013}
\communicated{S. Sarkar}

\begin{abstract}
Online multi-object tracking (MOT) is a longstanding task for computer vision and intelligent vehicle platform. 
At present, the main paradigm is tracking-by-detection, and the main difficulty of this paradigm is how to associate current candidate {detections} with historical tracklets. 
However, in the MOT scenarios, each historical tracklet is composed of an object sequence, while each candidate detection is just a flat image, which lacks temporal features of the object sequence. 
The feature difference between current candidate {detections} and historical tracklets makes the object association much harder. 
Therefore, we propose a Spatial-Temporal Mutual {Representation} Learning (STURE) approach which learns spatial-temporal representations between current candidate {detections} and historical {sequences} in a mutual representation space. 
For historical trackelets, the detection learning network is forced to match the representations of sequence learning network in a mutual representation space. 
The proposed approach is capable of extracting more distinguishing detection and sequence representations by using various designed losses in object association. 
As a result, spatial-temporal feature is learned mutually to reinforce the current detection features, and the feature difference can be relieved. 
To prove the robustness of the STURE, it is applied to the public MOT challenge benchmarks and performs well compared with various state-of-the-art online MOT trackers based on identity-preserving metrics.
\end{abstract}


\end{frontmatter}


\section{Introduction}
\label{sec1}
Online multi-object tracking (MOT) has become a critical scientific issue and perception technique required in many real-time applications such as intelligent driving~\citep{geiger2012are} and action recognition~\citep{RN583}. 
Considering the complexity of real-world scenarios, the MOT task is inherently complicated, and the key difficulties in MOT are appearance variations of the same target, similar appearances of different targets or frequent occlusion by a cluster of {objects}. 
With the development of deep learning, MOT has made extraordinary progress~\citep{RN550,wen2020ua-detrac}. 
MOT algorithms generate consistent trajectories by localizing and identifying multiple {objects} in consecutive frames, and they can be divided into two categories: offline and online MOT approaches. 
Offline approaches exploit the entire video information to extract tracking results. 
However, these algorithms are {inappropriate} for online applications such as intelligent driving. 
On the contrary, online MOT approaches only use the accessible data at the current moment.

\begin{figure}[!ht]
	\centering
	\includegraphics[width=1.0\linewidth]{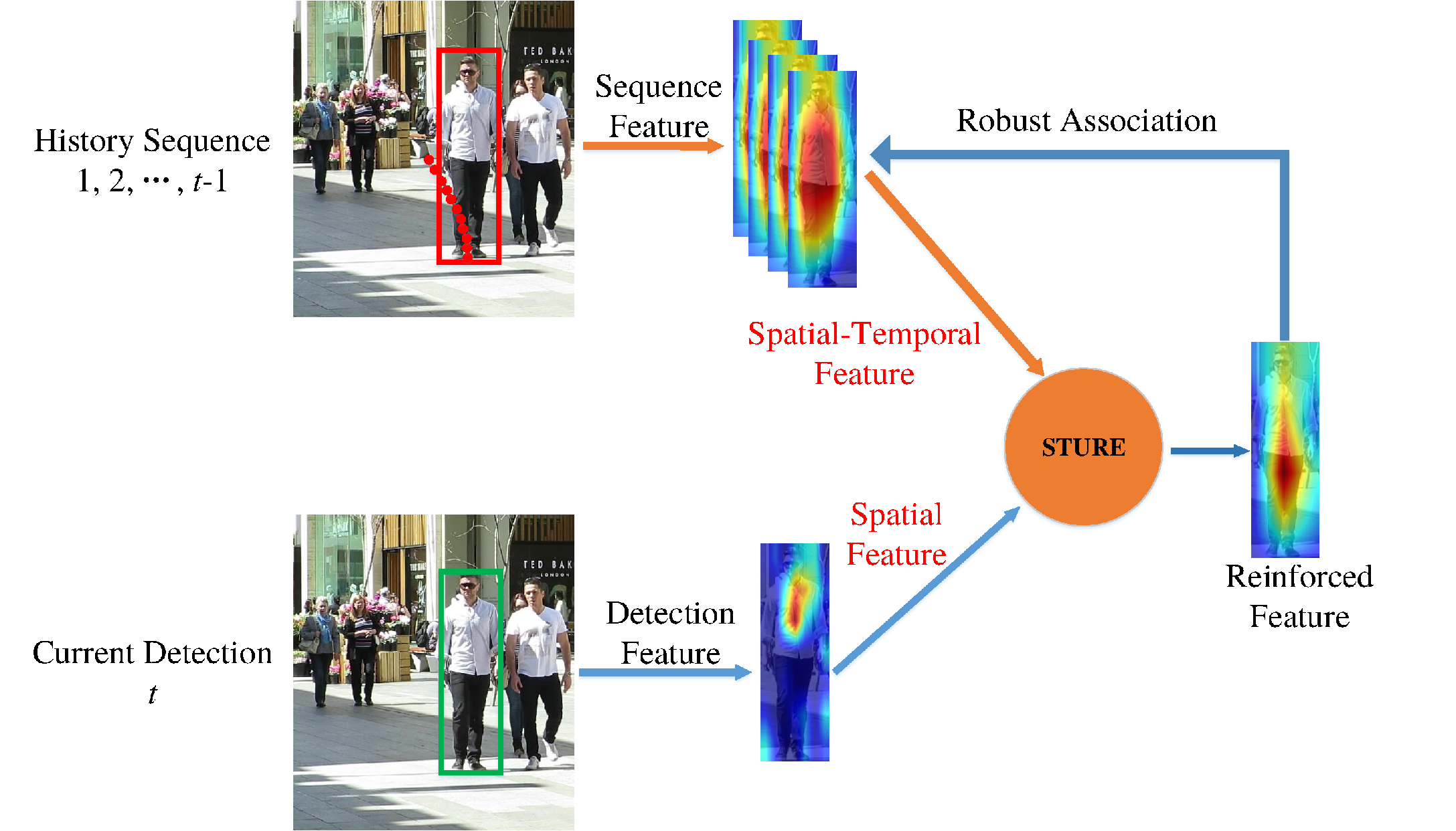}
	\caption{Spatial-temporal mutual representation learning and robust object association. 
		The spatial feature of the current detection is lack of temporal features {compared with} history sequence,
		and it merely highlights certain areas partly.
		Benefited from the STURE approach, temporal feature of historical sequences is learned to reinforce the feature of current detection,
		and the reinforced detection feature pays close attention to salient object further.}
	\label{fig:introduction}
\end{figure}

Due to the improvement of detection accuracy~\citep{ren2017faster}, the object association method which associates current detection results with historical {sequence} has been widely used. 
But the object association algorithm depends on perfect detection results.
Once the detection results of tracking objects are lost, ignored, or inaccurate, the process of tracking is likely to fail. 
These issues can be relieved by using recent high-precision single object trackers~\citep{RN1215,RN1212,RN1214} following a tracking-by-prediction paradigm. 
The single object trackers only employ the detections of the first frame to predict the status of a tracked object in the latter sequence frames~\citep{li2016robust}. 
Nevertheless, when the tracked object is occluded, these tracking methods will drift~\citep{RN717,2021Person}. 
To make up for the shortcomings of the tracking-by-prediction paradigm, we propose a method which combines the benefits of object association methods with single object trackers to improve the performance of MOT. 
A series of single object trackers are employed to track every object in the majority of video frames. 
When the score of tracking is lower than the threshold, the target association method will be used.

In addition, the object association in MOT refers to associating the current candidate detection with a series of historical tracklets~\citep{RN994,RN724}.
The historical tracklet usually contains a pedestrian sequence with temporal features, while the candidate detection only contains two-dimensional image representations. 
So the object association must be conducted between two different {modalities}, which are detection results and image sequences, respectively.
A number of studies~\citep{RN969,RN970} have demonstrated that extracting temporal features from image sequences can get more robust pedestrian sequences features in various complex environments. 
Nevertheless, such approaches neglect that the current detection image lacks temporal features during object association, so the current detection representation can not utilize the temporal features (\Cref{fig:introduction}). 
Meanwhile, the feature difference between current candidate {detections} and historical tracklets makes it more difficult to evaluate the similarity in object association. 
Therefore, it is particularly urgent and important to design an approach to reinforce current detection feature by using historical temporal knowledge.

To deal with the problems of the neglected temporal feature of current detection result and the feature difference in object association, we introduce a fancy Spatial-Temporal Mutual Representation Learning (STURE). 
The idea is enlightened by mutual learning strategy~\citep{RN983}, which learns feature collaboratively throughout the training process. 
{
In our proposed STURE, the temporal feature extracted a sequence learning network is transfered to the detection learning network.}
At the train stage, given a pedestrian sequence, the current detection features learned by the detection learning network are forced to fit the feature of the sequence learning network. 
By using STURE, these spatial-temporal features are learned mutually by the sequence learning network and the detection learning network. 
At the test stage, the reinforced detection learning network is used to extract the features of the current detection. 
Because of the learned temporal information, the enforced current detection features are robust to various complex environments just as the historical sequence features in \Cref{fig:introduction}.
At the same time, the feature difference problem between detection and sequence is solved, so the current candidate detection can be associated with the historical tracklets better in object association.

To sum up, our main contributions are listed as follows: 

\begin{itemize} 
	\item A superior STURE architecture is proposed to solve the problem of feature difference between the spatial features of current detection and the spatial-temporal features of historical sequence for object association.
	\item In order to enhance the mutual learning and identification ability of the proposed method, we have designed three loss functions: cross loss, modality loss and similarity loss, which will help the detection learning network obtain the temporal features. 
	\item A tracking-by-prediction MOT tracking paradigm is designed to alleviate the drift problem of single object tracking by using the reinforced features with STURE, which will improve the accuracy and robustness of the proposed method. 
	\item Abundant experiments performed on MOT benchmark with ablation studies are conducted, and these demonstrate that the proposed algorithm can obtain competitive tracking accuracy against the state-of-the-art online MOT methods.
\end{itemize}

\section{Related work}

\subsection{Multi-Object Tracking}
The current MOT solutions typically involve object association, which usually follow {a} tracking-by-detection paradigm. 
For example, {they} associate the detections across frames by calculating the pairwise affinity. 
Depending on whether the whole video information is employed, the MOT approaches are {divided} into online and offline approaches.
Offline methods~\citep{tang2017multiple} leverage both historical and future data. So they can take advantage of information about the entire video sequence.
They normally consider MOT as a graph optimization issue with different paradigms~\citep{li2020graph}, such as k-partite graph~\citep{dehghan2015gmmcp} and multicut~\citep{tang2017multiple}.

{
While online MOT approaches~\citep{chu2017online, RN1004, junbo2020a} don't utilize future information{, they} are likely to fail when the detection of the tracked target is inaccurate. 
Most of the previous approaches~\citep{bae2014robust,xu2019spatial-temporal} adopt a tracking-by-detection pipeline, whose performance is largely dependent on the detection results.
Recently, some~\citep{feng2019multi,RN455,RN542} combine the merits of single object tracking and data association to carry out online MOT and generally gain better tracking results.
FAMNet~\citep{RN542} integrates single object tracker and a object manager into a tracking model to recover false negatives. 
LSST~\citep{feng2019multi} utilizes a single object tracker to capture short term cues, a reidentification submodule to extract long term cues and a switcher-aware classifier to make matching decisions.
This strategy improves the tracking accuracy, but also brings a huge running cost.
MTP~\citep{kim2021discriminative} also considers discriminative appearance association and all previous tracks simultaneously to improve online tracking performance.}
In this study, a similarity association method with single object tracker and robust appearance is introduced to deal with the issue of imperfect detection.
It indicates that the new MOT approach can get commendable tracking results compared with existing algorithms.

\subsection{Spatial-Temporal Modeling}  
Extracting spatial-temporal feature is important for sequence frame information modeling.
Typically, recurrent structure is utilized to model temporal features~\citep{chung2017a, mclaughlin2016recurrent}.
The temporal average pooling~\citep{mclaughlin2016recurrent} is used at each time step to extract the feature of video sequences. 
Nevertheless, these methods can not deal with multiple adjacent regions. 
Lately, non-local neural network is utilized to model long-term temporal relations~\citep{RN580}.

For online MOT, a robust motion representation is important for object association. 
Recently, many online MOT methods~\citep{chu2017online,RN455,RN601} which use deep neural network have been formed. 
For example, siamese network~\citep{leal-taixe2016learning} has been applied to estimate the affinity of provided detections by aggregating targets appearance and optical flow information. 
In particular, Long Short-Term Memory (LSTM) is used to extract spatial representations~\citep{RN455}, and it processes detection results in the sequence one by one and outputs the similarity between them. 
In this study, we use {a} self attention mechanism to extract the temporal representations in pedestrian {sequences}.

\subsection{Mutual Learning} 
Generally, mutual learning~\citep{hinton2015distilling,romero2015fitnets,RN982} is a widely used cooperative learning method throughout training process. 
As for the learning methods, the process is performed by optimizing the KL divergence between the distribution of the final outputs from two different networks~\citep{RN983}.
Besides, the middle layer's output of the two networks is optimized~\citep{romero2015fitnets}. 
And the feature transferring is performed by calculating the similarities of cross-sample in metric learning problems~\citep{RN982, RN984}. 
In the study, temporal feature is learned mutually by optimizing the loss between current detections and historical sequence features in a mutual representation space.
Comparing to some works~\citep{RN982,RN927}, we design a new model architecture and loss functions.
In addition, the cross loss is designed on the basis of differences in the mutual representation space. 
The sequence learning network and the detection learning network are trained synchronously, rather than train the former first. 
To the best of our knowledge, in the online MOT methods, the proposed STURE is the first attempt to train a sequence learning network and a detection learning network in a mutual representation space.

\begin{figure*}[!ht]
	\centering
	\includegraphics[width=0.7\linewidth]{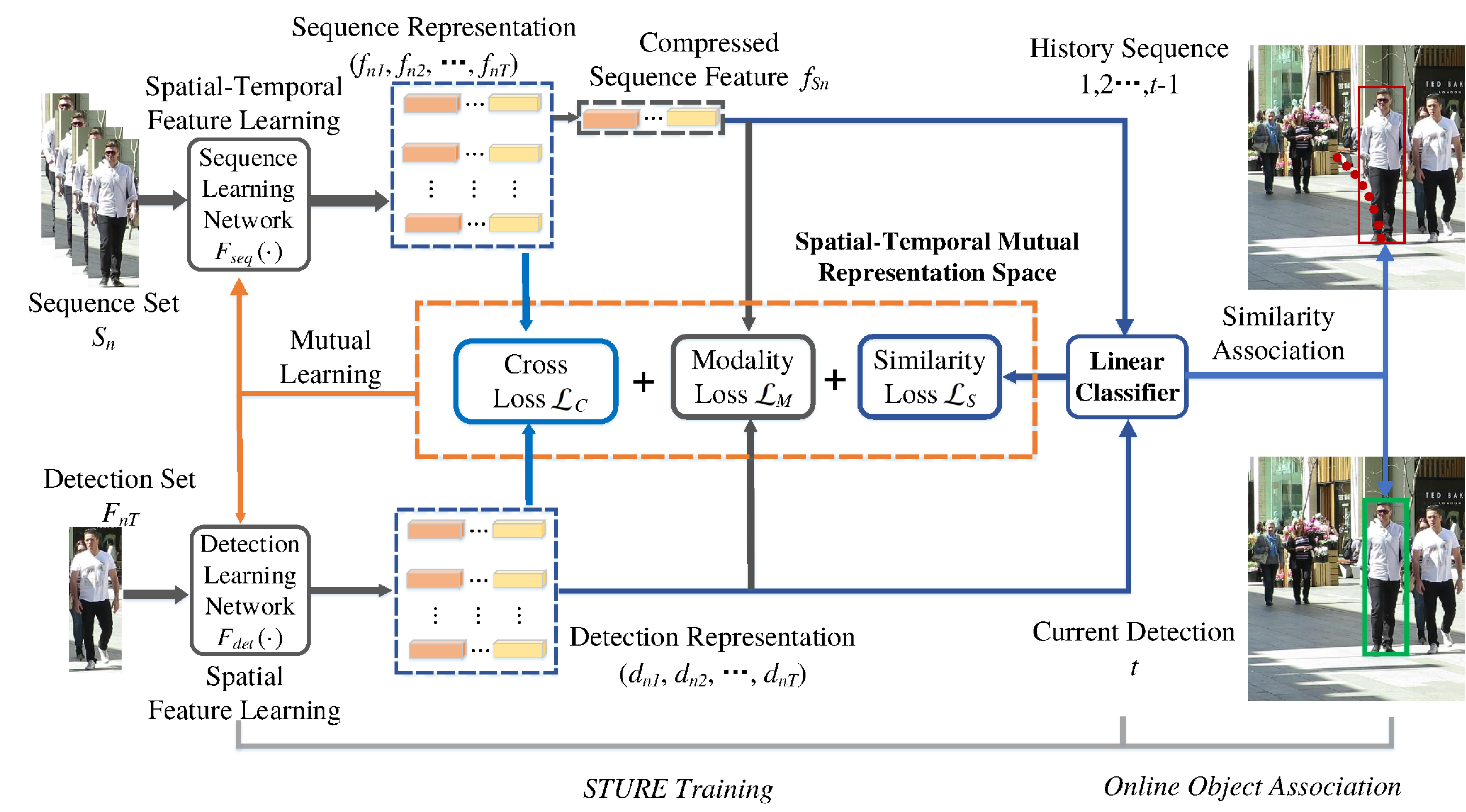}
	\caption{The architecture of the STURE approach in online MOT object association.
		Different colors denote different representation transfer paths.
		The blue path indicates the computation of cross loss $\mathcal{L}_C$ by representation difference between sequence and detection, 
		while the gray path represents the computation of modality loss $\mathcal{L}_M$ by mutual modality between detection sequence and single detection.
		Besides, the similarity loss $\mathcal{L}_S$ {is} used to train the sequence learning network $\mathcal{F}_{seq}(\cdot)$ and detection learning network $\mathcal{F}_{det}(\cdot)$.
		Specifically, the spatial-temporal feature are learned simultaneously.
		Finally, the online object association is performed according the similarity between the historical pedestrian sequence and the current detection.}
	\label{fig:network_architecture}
\end{figure*}

\section{Approach}
Firstly, we present a complete model architecture overview of STURE.
Secondly, we give the detailed information of the detection learning network and the sequence learning network. 
Thirdly, we introduce the design of loss function and the training method.
In the end, the trained models are utilized to associate current {detections} and historical tracklets in online MOT. 

\subsection{Architecture Overview}
The architecture of STURE in the detection-to-sequence object association is presented (see Figure \ref{fig:network_architecture}). 
The sequence learning network learns spatial features and deals with temporal relations between sequence frames synchronously; 
and the detection learning network learns spatial features of the current detection. 
The spatial-temporal features are extracted by sequence learning network and detection learning network in a mutual representation space with the designed losses. 
By optimizing the total objective function, the detection representations and the sequence representations will be learned simultaneously. 
In the end, the detection-to-sequence object association is conducted according the similarity between {current detections and historical tracklets}.  
The detailed procedures are presented in the following parts.

\subsection{Spatial-Temporal Modeling Network}
\vspace{5pt}
\noindent
\subsubsection{Sequence Learning Network}
\label{sec:videomodel}

In order to extract the spatial-temporal feature of historical pedestrian tracklets in a single model, Convolutional Neural Network (CNN) with non-local self attention mechanism~\citep{RN580} is used in the sequence learning network.
By exploiting temporal relations within the sequence, the non-local neural network can summarize image-wise features into a single feature, and then output the activation of each location by the weighted average of each position using the input representation~\citep{RN580}.

\begin{figure}[!ht]
	\centering
	\includegraphics[width=0.78\linewidth]{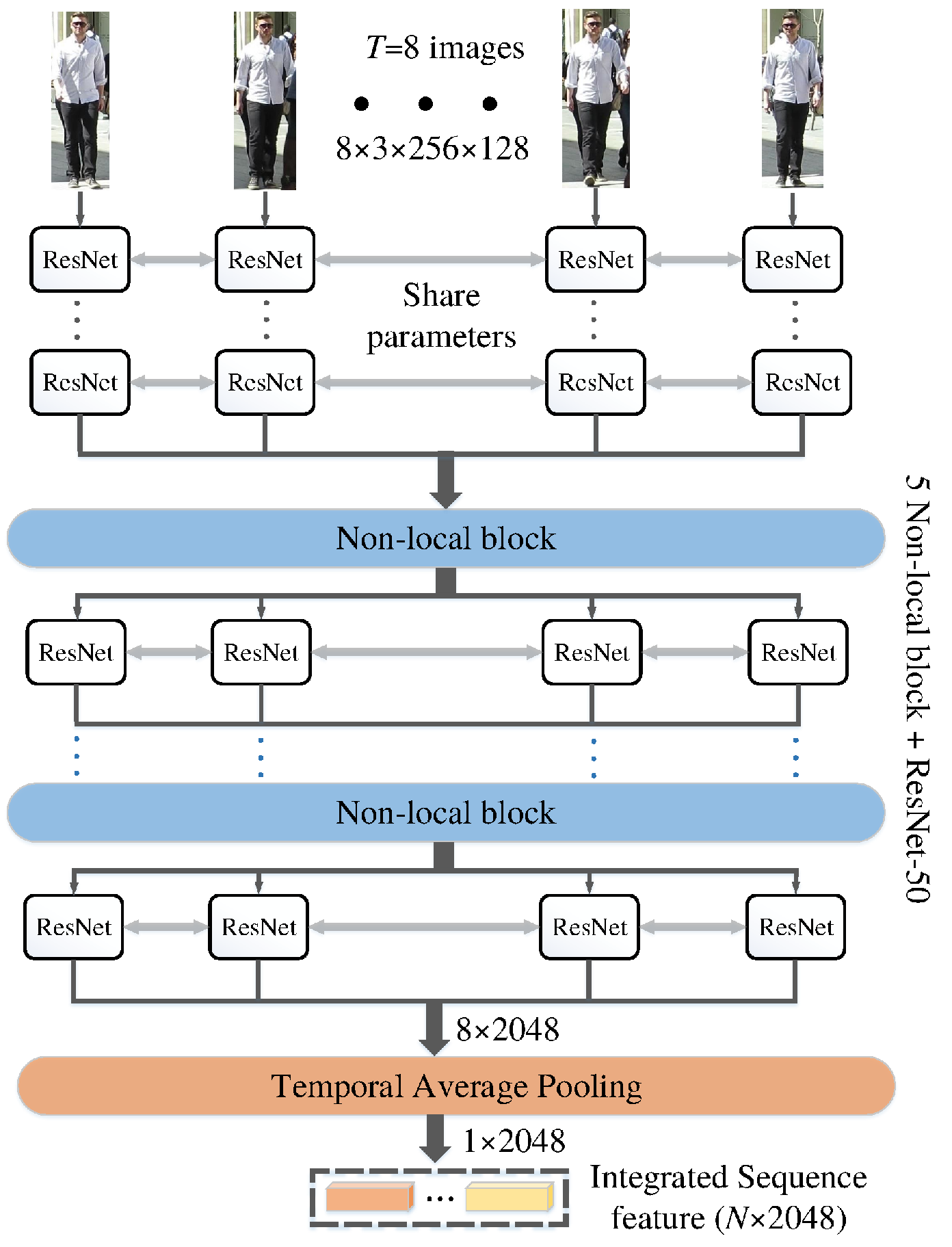}
	\caption{The architecture of the sequence learning network. 
		All 8 detection images constitute the input sequence, and each pedestrian is $256\times 128$, the output of sequence learning network is $1\times2048$.}
	\label{fig:sequence}
\end{figure}

\Cref{fig:sequence} indicates the architecture of the sequence learning model based on ResNet{~\citep{he2016deep}}.
Five non-local layers are particularly embedded into CNN, 
and the final down-sampling operation of the ResNet in the last convolution is deleted to obtain a high-resolution feature map~\citep{RN985}.
Given $N$ pedestrian sequences $\mathcal{S}=\{S_n\}^N_{n=1}$, which can be described as:
\begin{equation}\label{eq:pedestrian_sequence}
S_n=\{F_{n1},F_{n2},...,F_{nT}\},
\end{equation}
where $n$ is the index of a pedestrian sequence, $n=1,2,...,N$, and $T$ is the number of frames for each pedestrian sequence (here $T$ is 8). 
So the sequence feature can be extracted by $\mathcal{F}_{seq}(\cdot)$:
\begin{equation}\label{eq:sequence_network}
\left( f_{n1},f_{n2},...,f_{nT} \right)=\mathcal{F}_{seq}(S_n),
\end{equation}
where $f_{nt} \in R^{D}$ is the feature of frame $F_{nt}$ in sequence $S_{n}$, and $t=1,...,T$.
The sequence features of several pedestrians are compressed into a single sequence representation $f_{Sn}\in R^{D}$ by using 3D pooling ($TAP$):
\begin{equation}\label{eq:tap}
f_{Sn}=TAP\left( f_{n1},f_{n2},...,f_{nT} \right).
\end{equation}

\subsubsection{Detection Learning Network}
\label{sec:imagemodel}

In order to learn spatial feature, the detection learning network uses ResNet which has removed the last fully-connected layer.
The last downsampling step of ResNet-50 is deleted like the sequence learning network to get higher spatial resolution and more-refined detection representations.
Once the temporal features among pedestrian sequences are ignored, the input sequence $\mathcal{S}$ can be regarded as many independent detections $\{F_{nt}\}_{n=1,t=1}^{N,T}$.
So the detection learning network $\mathcal{F}_{det}(\cdot)$ is utilized to learn the representations of the independent detections: 
\begin{equation}\label{eq:detection_network}
d_{nt}=\mathcal{F}_{det}(F_{nt}),
\end{equation}
where $d_{nt}\in R^{D} (t=1,...,T$) is the corresponding detection feature of the sequence frame $F_{nt}$.

Both the architecture of sequence and detection learning models utilize ResNet as backbone,
and the only difference between them is that the former adds additional non-local neural network to extract temporal features.

\subsection{Spatial-Temporal Mutual {Representation} Learning}
\label{sec:stml}
Generally speaking, in online MOT, the result of object association has a lot to do with the robust feature learning.
Extracting temporal knowledge among pedestrian sequence {will improve} the robustness of the spatial features in various environmental challenges~\citep{RN986}.
Nevertheless, the detection learning network (its inputs are flat images) is unable to deal with temporal correlations, which {prevents} it from learning temporal features.
To deal with the issue, the proposed STURE makes the detection learning network's outputs to match the representations of the sequence learning network in a mutual representation space.

For the specific pedestrian sequence $S_n$, \Cref{eq:sequence_network}, \Cref{eq:tap} and \Cref{eq:detection_network} are utilized to extract sequence features $f_{Sn}$ and detection features $d_{nt}${, respectively}.
Because $\mathcal{F}_{seq}(\cdot)$ learns spatial features and temporal correlations among pedestrian sequences $F_{nt}$, $f_{Sn}$ contains both spatial and temporal features.
To extract sequence's temporal feature representation $f_{Sn}$ in the detection learning network and sequence learning network mutually, the STURE is designed to optimize the following losses.

\vspace{3pt}
\noindent
\subsubsection{Cross Loss}
The STURE enforces the detection learning network to match more refined temporal features in mutual representation space. 
In the circumstances, the STURE is formed to optimize the error between pedestrian sequence representations and the corresponding detection representations.
It learns temporal {features} in the sequence learning network and the detection learning network mutually to utilize deep feature.
The representation of the {tracked} object is expressed by cross-sample.
The feature of the whole sequences can be described as $\{f_{nt}\}_{n=1,t=1}^{N,T}$ .
We use cross-sample to measure the difference between detection-detection, detection-sequence and sequence-sequence.
The Euclidean difference matrix of the cross-sample $M^{seq}\in R^{NT\times NT}$ {can be indicated as:}

\begin{equation}\label{eq:m_seq}
M^{seq}={
	\left[ \begin{array}{cccc}
	m_{11} & m_{12} & ... & m_{1N}\\
	m_{21} & m_{22} & ... & m_{2N}\\
	... & ... & ... & ... \\
	m_{N1} & m_{N2} & ... & m_{NN}\\
	\end{array} 
	\right ]},
\end{equation}
where each submatrix $m_{ij}$ is a $T \times T$ matrix with same value, $i,j=1,2,...,N$. 
In this, $T$ is the length of pedestrian sequence and $N$ is the number of sequences.
\begin{equation}\label{eq:m_ij}
m_{ij}={
	\left[ \begin{array}{cccc}
	s_{ij} & s_{ij} & ... & s_{ij}\\
	s_{ij} & s_{ij} & ... & s_{ij}\\
	... & ... & ... & ... \\
	s_{ij} & s_{ij} & ... & s_{ij}\\
	\end{array} 
	\right ]},
\end{equation}
where $s_{ij}$ is the difference submatrix between the sequence $i$ and $j$ ($i,j=1,2,...,N$).

The cross detection differences $M^{det}\in R^{NT\times NT}$ are forced to fit the cross sequence frame difference matrix $M^{det}$ to learn detection {features} in the mutual representation space.
\begin{equation}\label{eq:m_det}
M^{det}={
	\left[ \begin{array}{cccc}
	c_{11} & c_{12} & ... & c_{1H}\\
	c_{21} & c_{22} & ... & c_{2H}\\
	... & ... & ... & ... \\
	c_{G1} & c_{G2} & ... & c_{GH}\\
	\end{array} 
	\right ]},
\end{equation}
where $G=H=N \times T$, and every element $c_{gh}$ is the Euclidean difference between the detection $g$ and detection $h$ ($g,h=1,2,...,NT$).

{
In this way, the temporal feature can be transfered to detection learning network.}
The cross loss is formulated as:
\begin{equation}\label{eq:m_cross}
\mathcal{L}_C=\frac{1}{G H} \sqrt{ \sum_{g=1}^{G} \sum_{h=1}^{H} | M_{gh}^{seq}-M_{gh}^{det} |^2 },
\end{equation}
where $\mathcal{L}_C$ denotes the error between the cross difference matrix of the detection and sequence.
We can use \Cref{eq:m_cross} to reconstruct the detection representation function $\mathcal{F}_{det}(\cdot)$ by learning the sequence representations ${(F_{nt},f_{nt})}_{n=1,t=1}^{N,T}$,
and it can be viewed as a continuous representation reconstruction from a group of data~\citep{RN987,RN988}.
The sequence and detection learning networks are similar to FitNets~\citep{romero2015fitnets}, except the output of the model. 
They are converted to the identical size with an additional convolution.
By comparison, we don't require extra convolution because the outputs of the sequence and detection learning networks have identical size.
After training, the detection learning network will be able to learn from the pedestrian sequences, which can make it gain desired temporal features.

Except for cross-sample loss, other identification losses are added to extract discriminative representations in detection-to-sequence object association.
Any loss that can improve the discriminability is feasible in the same way.
Particularly, in our work, the modality loss and the similarity loss are used.

\vspace{3pt}
\noindent
\subsubsection{Modality Loss}
The triplet loss~\citep{cheng2016person} is utilized to keep interindividual differences in the mutual representation space.
In this study, two types of modality differences are designed, including cross-modality and within-modality loss.

The cross-modality loss keeps the difference between detection and sequence representation, and it is able to enhance the representation discriminability of different modalities. 
It is formulated as:
\begin{equation}\label{eq:I2Sloss}
\begin{aligned}
\mathcal{L}_{cross} = \max (0, \max_{d_p\in S_b^+} E(s_b, d_p) - \min_{d_n\in S_b^-} E(s_b, d_n) + m) \\
+ \max(0, \max_{s_p\in S_b^+} E(d_b, s_p) - \min_{s_n\in S_b^-} E(d_b, s_n) + m ),
\end{aligned}
\end{equation}
where the previous term is the sequence-to-detection loss, and the latter is the detection-to-sequence loss.
$m$ indicates the pre-set margin, $E(\cdot,\cdot)$ means the difference in Euclidean space.
$S_b^+$ and $S_b^-$ are the positive and negative sample datasets of the pedestrian ($s_b$ and $d_b$), respectively. 

Similarly, the within-modality loss keeps relative differences within a same modality, which makes the method discriminate the finely grained features to various objects in the same modality. 
It is formulated as:
\begin{equation}\label{eq:I2Iloss}
\begin{aligned}
\mathcal{L}_{within}= \max(0, \max_{s_p\in S_b^+} E(s_b, s_p) - \min_{s_n\in S_b^-} E(s_b, s_n) + m ) + \\
\max (0, \max_{d_p\in S_b^+} E(d_b, d_p) - \min_{d_n\in S_b^-} E(d_b, d_n) + m ) ,
\end{aligned}
\end{equation}
where the previous term is the sequence-to-sequence loss, and the latter is the detection-to-detection loss.

The losses of the within-modality and the cross-modality are able to extract detection-to-sequence feature more efficiently.
So we integrate the two kinds of modality losses: within-modality and cross-modality losses.
The final modality loss $\mathcal{L}_{M}$ can defined as:
\begin{equation}\label{eq:tripletloss}
\mathcal{L}_{M}= \mathcal{L}_{cross}+\mathcal{L}_{within}.
\end{equation}

\vspace{3pt}
\noindent
\subsubsection{Similarity Loss} 
Because pedestrian identities are category-level information provided in MOT datasets, two same weight classifiers are constructed to convert the detection representations and sequence representations to a mutual representation space.
Several fully-connected layers followed by a softmax function constitute the object classifier; 
and the number of output channels is equal to the number of identities in MOT datasets.
So the similarity loss $\mathcal{L}_{S}$ is formed as the cross entropy loss between the inferred object label and the ground truth label.
\begin{equation}\label{eq:cross_entropy}
\mathcal{L}_{S}=- \sum_{i} y_{i}^{'}log(y_i),
\end{equation}
where $y_{i}^{'}$ is the ground truth, and $y_{i}$ is the predicted label.

\vspace{5pt}
\noindent
\subsubsection{Total Loss}
The spatial-temporal features are learned simultaneously in detection learning network and sequence learning network.
So the total loss is composed of cross loss, modality loss and similarity loss as follows:
\begin{equation}\label{eq:loss}
\mathcal{L}=
\mathcal{L}_{C} + 
\mathcal{L}_{M} + \mathcal{L}_{S}.
\end{equation}

\subsection{Training Strategy}
\label{sec:training_strategy}
The labeled bounding boxes and identity data of pedestrians supplied in the MOT training dataset are used to produce pedestrian {sequences} and current candidate detections.
We utilize it to train our proposed model. 

\vspace{5pt}
\noindent
\subsubsection{Data Augmentation}
The training set in MOT datasets does not contains enough pedestrian tracklets, and each pedestrian sequence contains limited detection results. 
Therefore, the association model will be liable to underfit for the training data. 
In this study, a number of image augmentation methods are employed to relieve these difficulties.

Firstly, the training set is augmented by cropping and rescaling the input detection images randomly;
and horizontal flip is also utilized.
Furthermore, in order to simulate the noise environment for tracking and improve the robustness of the proposed model,
some noise data is mixed to the pedestrian sequences by substituting detections of other pedestrians randomly.
While some sequences may contain only a small number of pedestrian images in the training set, 
each tracklet is sampled with the same probability to relieve the sample disequilibrium problem,
and the appropriate channel number is prepared for the sequence learning network.

\vspace{5pt}
\noindent
\subsubsection{Data Sampling}
\label{sec:sampling}
In order to optimize the proposed network by various target functions in \Cref{eq:loss}, an especial data sampling method is utilized in MOT datasets.
$P$ pedestrians are selected stochastically in every training iteration.
$Q$ sequences are generated randomly for every pedestrian, 
and each sequence contains $T$ detections. 
If a sequence is shorter than $T$, {it} will be sampled by equal probability from the preserved frames to meet the requirement of model input.
We input the whole $N=P\times Q$ pedestrian sequences to the sequence learning network.
Besides, no more than $M$ recent tracking results of the object is preserved.
At the same time, $N\times T$ current candidate detections which constitute a detection batch are fed to the detection learning network.
To reduce the computational cost, all data for each detection batch is reused to evaluate three different target functions in \Cref{eq:loss}.

\vspace{5pt}
\noindent
\subsubsection{Selective Back-propagation}
For each input data, the goal of STURE is to force the sequence learning network and detection learning network to output similar features.
It's easy to find that the two networks will have the same representations if an optimal solution is used to minimize the STURE loss.
Hence, updating the sequence learning network by cross loss will restraint temporal feature learning.
In this case, the detection learning network won't learn the desired temporal feature.
To solve this problem, in this study, the cross loss $\mathcal{L}_{C}$ is not used to update the sequence learning network at the training stage.
So the selective back-propagation strategy makes the detection learning network learn more robust representation, 
and it won't diminish the ability of learning temporal features from the sequence learning network.

\vspace{5pt}
\noindent
\subsubsection{Similarity Computing}
{
To evaluate the similarity between candidate detections and the historical tracklets, we merge the integrated sequence feature $f_{Sn}$ (1$\times$2048) and the output $d_{nt}$ (1$\times$2048) of detection learning network into one representation, 
and then put it into the linear classifier, which evaluates the similarity between them.
The linear classifier has three fully connected layers and its input dimensions are 4096, 256, and 32, respectively.}
Meanwhile, each fully-connected layer incorporates a batch normalization layer and an activation function.
The last layer of the linear classifier will output the affinity between candidate detections and historical tracklets. 
Afterwards, a softmax operation is followed,
and the similarity loss $L_S$ is formed by the cross entropy loss between the inferred label and the ground truth.

Finally, the whole network is trained by optimizing the total loss according to \Cref{eq:loss}.

\subsection{Object Association}
\label{sec:i2vtesting}
A single object tracker is utilized to track the object in each {video} frame.
Once the single object tracking procedure turns into drifting, it will be suspended and the status of the tracked object will change to be drifting.
So the tracked object status can be described as:
\begin{equation}
\ track \ status=\left\{
\begin{array}{rcl}
tracked & {if \ s > \tau_s \ and \ o_{a} > \tau_o}\\
drifting & {otherwise}, 
\end{array} \right.
\end{equation}
where $s$, $\tau_s$, and $\tau_o$ are tracking score, track score threshold, and overlap rate between the tracked target and detection, respectively.  
The average overlap of the pedestrian sequences $ o_{a} $ is denoted as 
\begin{equation}\label{eq:overlap_mean}
o_{a}=\frac{1}{I} \sum_{i=1}^{I} o\left(t_i,D_I\right).
\end{equation}
When determining the status of tracked object, the average value of $ o_a $ in the historical $ I $ frames is taken into consideration. 
For \Cref{eq:overlap_mean}, the overlap rate between current detection result and the tracked object is indicated as{:} 
\begin{equation}
\label{overlap_target_detection}
o \left(t_i,D_I\right) =\left\{
\begin{array}{rcl}
0& {if \ \tau_o > max \left(IoU \left(t_i,D_i\right) \right) } \\
1& {otherwise}, \\
\end{array} \right.
\end{equation}
where $ o \left(t_i,D_I\right) $ is assigned to 0 when the maximal intersection over union (IoU) of the previous tracked target $ t_i \in T_i $ (all tracking object within $i$ frames) and the current detection $D_i$ in $i$ video frames is lower than $\tau_o$. 
Otherwise, $ o \left(t_i,D_i\right) $ is assigned to 1.

The motion information is utilized to choose current candidate detection results before evaluating the affinity for similarity association.
Once the tracked object drifts, the size of the pedestrian bounding box at the frame $ t-1 $ will remain unchanged, 
and a linear prediction method is utilized to infer the object's position at the latest moment $t$. 
Let $ p_{t-1}=\left[x_{t-1},y_{t-1}\right] $ be the center position of the tracked object at frame $ t-1 $,
so the velocity $ v_{t-1} $ of the tracked object at frame $ t-1 $ is evaluated as{:}
\begin{equation}
v_{t-1}=\frac{1}{L} \left( {p_{t-1}-p_{t-L}} \right),
\end{equation}
where $ L $ indicates the length of historical sequence. 
Hence, the position of the tracked object in frame $t$ is inferred as{:}
\begin{equation}
p_t=p_{t-1}+v_{t-1}.
\end{equation}

If the detection result which around the predicted position of the tracked object is not overlapped with any other tracked object (the difference between inferred position and detection result is smaller than a threshold value $ \tau _d $), it will be viewed as a candidate detection in the current frame $t$. 
The affinity is measured between current detections and historical sequences. 

\vspace{5pt}
\noindent
\subsubsection{Similarity Association}
\begin{figure}[!ht]
	\centering
	\includegraphics[width=1.0\linewidth]{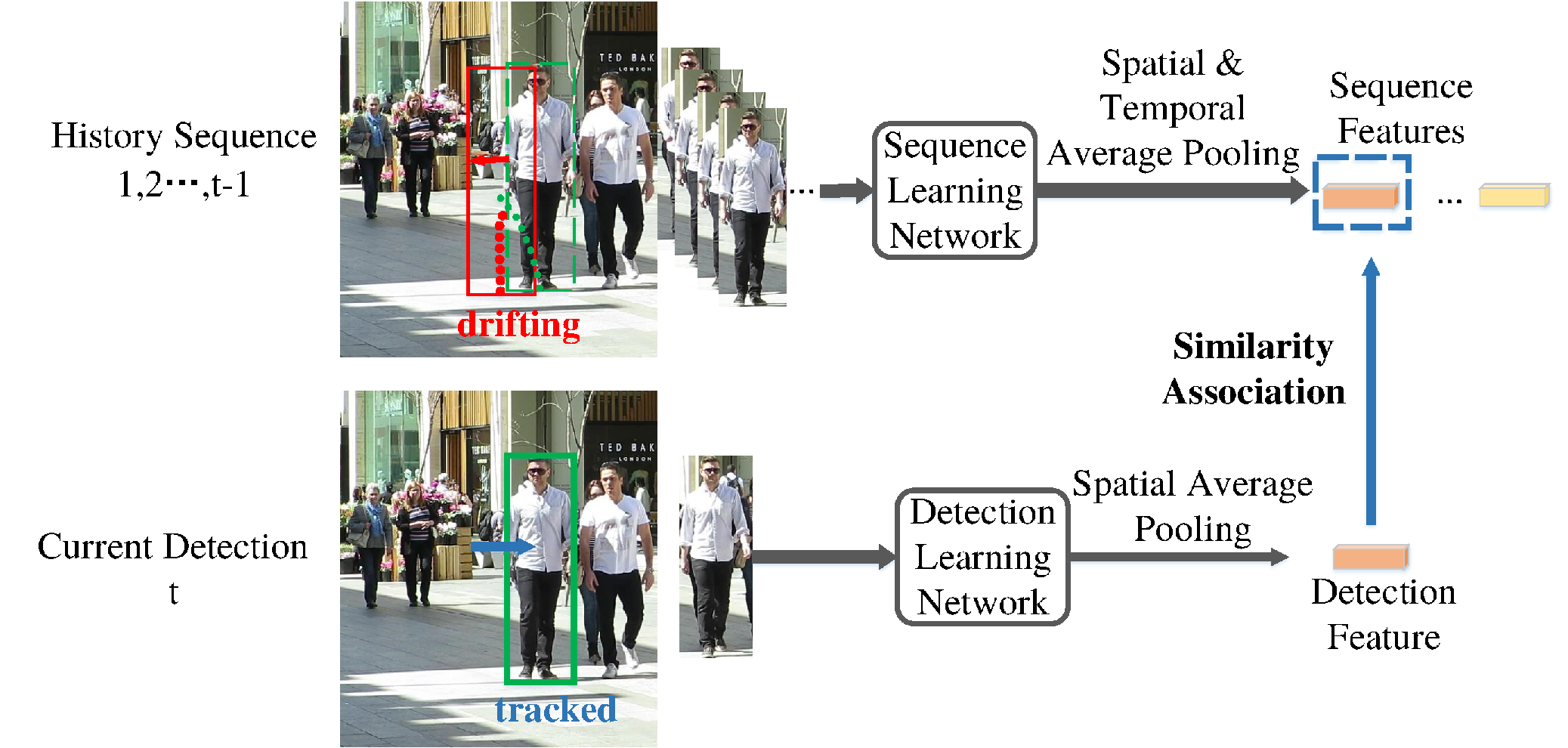}
	\caption{The pipeline of detection-sequence object association. 
		When the single object tracking procedure turns into unreliable, tracked target is assigned to drifting status and conduct detection-to-sequence object association according the similarity score between the historical target sequence and the current detection.}
	\label{fig:association}
\end{figure}

The similarity association is utilized, as shown in Figure \ref{fig:association}, to decide whether the status of tracked object should be converted to tracked or kept drifting.
In this stage, each association operation is performed between flat images and a mass of pedestrian sequences.
In the process of detection-to-sequence association, the current detection and the historical tracklet representations are extracted by the detection learning network and the sequence learning network, respectively.
After extracting representations, the similarity between current detection and historical tracklets is evaluated.
Then the detection-to-sequence object association is conducted according to the similarity.
The most similar detection will be selected and a similarity threshold value $ \tau_a $ is set to determine whether the drifting object is link to the sequence. 

In this tracking method, it is a natural solution that we use the highest tracking score of the target in the confidence map to evaluate the reliability of a single object tracker.
Nevertheless, if we only use the tracking score, the false alarm detections with high confidence value are liable to be tracked persistently.
In the ordinary way, the tracked object which does not overlap with other detection objects continuously is much more likely to be false positive.
To solve this problem, the single object tracker and the overlap rate of bounding boxes and are used to remove the false positives.

Finally, the drifting objects and current detection results are assigned, which based on the paired affinity values between the historical sequences and the candidate detection results.

\vspace{5pt}
\noindent
\subsubsection{Object Appearing and Disappearing}
In the process of MOT, using the detection results supplied by MOT benchmark~\citep{RN583}, we can initialize a newly emerged object and start the tracking process.
When the overlap rates of a current detection result with all tracked objects are lower than a threshold, it will be viewed as a new potential object.
To prevent false alarm detection results, when a pedestrian sequence in the new candidate detection is greater than a threshold value $\tau_i$ during $L$ frames continuously, it will be viewed as an initial tracked object.

With respect to object disappearing, a tracked target which does not overlap with any other detections will be viewed as drifting and then will be deleted from the tracked list.
The process of single object tracking will be terminated when it maintains drift status over than $ \tau _t$ frames or directly moves out of view.

\section{Experiments}
\subsection{Datasets and Evaluation Protocol}
\subsubsection{Benchmark Datasets}
The proposed method is evaluated on MOT16~\citep{RN583}, MOT17{~\citep{RN583} and MOT20~\citep{2019CVPR19}} datasets.
In total, there are 7 fully labeled training videos and 7 testing videos recorded by static or moving cameras in MOT16. 
The MOT17 has as many video sequences as the MOT16, while it supplys three extra image-based object detectors: DPM~\citep{RN584}, Faster-RCNN~\citep{ren2017faster}, and SDP~\citep{yang2016exploit}, which have different detection accuracy and noise levels, and they will support the test of different MOT methods.
The MOT20 consists of 8 new sequences depicting very crowded challenging scenes. 

\vspace{5pt}
\noindent
\subsubsection{Evaluation Protocols}
Various evaluation protocols of the MOT benchmarks~\citep{RN583} are utilized for a fair comparison. 
Except for the classic multi-object tracking accuracy (MOTA)~\citep{RN475} and multi-object tracking precision (MOTP)~\citep{RN550}, the evaluation metrics also contain the ratio of correctly identified detection (IDF1), ID recall~\citep{ristani2016performance} (IDR, the fraction of ground-truth detections that are correctly identified), the total number of false positives (FP), mostly tracked targets (MT), mostly lost targets (ML), the total number of identity switches (IDS), the total number of times a trajectory is fragmented (Frag), and the processing speed (Hz). 
Particularly, ID recall is added by~\citep{ristani2016performance} and has been introduced to the MOT benchmarks. 
It can be used to evaluate the consistency of the predicted identities with the actual identities.

\subsection{Implementation Details}
First of all, we use the ECO~\citep{danelljan2017eco} as the single object tracker in our proposed method.
The ResNet pre-trained on the ImageNet~\citep{RN991} is exploited as the backbone module, and the approach in~\citep{RN580} is adopted to initialize the parameters of the non-local layers. 
As for ResNet-50, the length of its output $D$ is $2048$.
The maximum preserved results in the trajectory $M$ is assigned to 100 and the length of tracklet $T$ is assigned to 8.
Every detection result of the tracked pedestrian is resized to 256 $\times$ 128. 
The batch size is assigned to 32.
The Adaptive Moment Estimation (Adam)~\citep{RN586} optimizer with the learning rate of $10^{-4}$ is used to optimize the proposed model.

The values of tracking {parameter} are assigned on the basis of the MOTA results.
Given $ F $ as the raw frame frequency, 
the initialization threshold value of trajectory $ \tau _i $ is assigned to $ 0.2F $. 
The termination threshold value of trajectory $ \tau _t $ is assigned to $ 2F $. 
The distance $ L $ for evaluating whether the object is tracked is assigned to $ 0.3F $.
The thresholds of the appearance similarity $ \tau _a $ and tracking score $ \tau _s $ are assigned to $ 0.8 $ and $ 0.2 $, respectively. 
The threshold values of the difference $\tau_d$ and overlap $\tau_o$ are $2$ and $0.5$. 
Besides, the threshold values of the tracking score and appearance are selected by grid search.
The experiment device we used is a workstation with an Intel Core i9-9820X CPU.
{The} MOT algorithm is implemented with Python by the Pytorch 1.3.0 library~\citep{paszke2019pytorch} and it is run in the Linux environment of Ubuntu 18.04.
The whole training procedure takes 3 hours for 80 epochs on a NVIDIA GeForce RTX 2080Ti.

During test stage, the representations of a detection result are extracted by the detection learning network. 
Firstly, for every whole pedestrian sequence, it is split into many 32-frame sequences.
The sequence learning model is used to extract sequence representation in every pedestrian sequence. 
The last compressed sequence representation is the average of the whole sequence features.

\subsection{Evaluations on {MOT benchmark}} 

{
The designed STURE is compared with the various types of MOT approaches.
The evaluations are shown in \Cref{tab:performance_MOT16}, \Cref{tab:performance_MOT17} and \Cref{tab:performance_MOT20}{,} respectively.
N1T~\citep{baisa2019development} is the approaches of hand-crafted representation, 
and PHD\_DAL~\citep{2019Online}, HISP~\citep{baisa2021robust}, GMPHD\_ReId~\citep{baisa2021occlusion} and GNN~\citep{li2020graph} are the methods based on deep representation.
It is obvious that those approaches which use deep {representation} can perform better than the traditional approaches.
Thus, to a certain extent, the proposed approach gains excellent performance compared with the published algorithms based on deep learning. 
Moreover, compared with discriminative appearance association in DDAL~\citep{RN601}, MTP~\citep{kim2021discriminative} and LSST~\citep{feng2019multi}, our method also have superior online tracking performance.
SORT~\citep{bewley2016simple}, IOU\_KMM~\citep{urbann2021online}, Tracktor++~\citep{bergmann2019tracking} and FlowTracker~\citep{nishimura2021sdof} try to combine detection into tracking to improve running speed without losing too much precision. 
}

{
The proposed online MOT algorithm STURE is tested on the test sets of MOT16, MOT17 and MOT20 benchmarks, and it has been compared with other online methods. 
\Cref{tab:performance_MOT16}, \Cref{tab:performance_MOT17} and \Cref{tab:performance_MOT20} indicate the tracking results on MOT16, MOT17 and MOT20 benchmark datasets{,} respectively. 
The proposed STURE gains better MOTA score and is compared with the other approaches with respect to MOTA, MOTP, IDF1, IDR, FP, FN, MT, ML, IDS, and Frag. 
Compared with the second best existing online methods, for MOT16, STURE has gain best performance in MOTA and Frag.
For MOT17, STURE has best performance in MOTA and FP.
And particularly for MOT20, STURE has best performance in MOTA and IDF1.
We can see that STURE has gain a good performance in both precision and speed against other online tracking methods in various metrics, which demonstrates the advantages in MOT.
}

\begin{table*}[]
\centering
\caption{Evaluations on the MOT16 benchmark from the existing online MOT algorithms.
	$\uparrow$ means the larger the better and $\downarrow$ means the smaller the better.
	The optimal results are shown in \bfseries{bold}.}
\renewcommand\arraystretch{1.4}
\begin{center}
	\setlength{\tabcolsep}{1.9mm}{
		\begin{tabular}{l |c |c | c | c | c | c |c  |c | c | c}
			\hline 
			Methods  
			&\multicolumn{1}{|c|}{MOTA\,$\uparrow$} &\multicolumn{1}{|c|}{MOTP\,$\uparrow$}
			&\multicolumn{1}{|c|}{IDF1\,$\uparrow$}
			&\multicolumn{1}{|c|}{IDR\,$\uparrow$} &\multicolumn{1}{|c|}{FP\,$\downarrow$} 
			&\multicolumn{1}{|c|}{MT\,$\uparrow$}
			&\multicolumn{1}{|c|}{ML\,$\downarrow$}
			&\multicolumn{1}{|c|}{IDS\,$\downarrow$} &\multicolumn{1}{|c}{Frag\,$\downarrow$}
			&\multicolumn{1}{|c}{{Hz}\,$\uparrow$} \\
			\cline{1-11}
			N1T~\citep{baisa2019development}    &33.3 &\bfseries76.9 &25.5 &36.1 &\bfseries1,750 &5.5 &56.0 &3,499 &3,594 & {9.9} \\
			PHD\_DAL~\citep{2019Online}   &35.4 &75.8 &26.6 &38.6 &2,350 &7.0 &51.4 &4,047 &5,338 &{3.5} \\
			HISP~\citep{baisa2021robust}   &37.4 &78.7 &30.5 &40.3 &3,222 &7.6 &50.9 &2,101 &2,151 &{3.3} \\
			GMPHD\_ReId~\citep{baisa2021occlusion}   &40.4 &73.7 &\bfseries50.1 &44.5 &6,569 &11.5 &\bfseries43.1 &789 &2,519 &\bfseries {31.6}\\
			DDAL~\citep{RN601}     &43.9 &74.7 &45.1 &34.1 &6,450 &10.7 &44.4 &\bfseries676 &1,795 &{0.5} \\
			OTCD~\citep{2019Real}  &44.4 &72.2 &45.6 &\bfseries47.9 &5,759 &11.6 &47.6 &759 &1,789 &{17.6}\\
			Ours     &\bfseries44.6 &74.9 &39.9 &44.3 &3,543 &\bfseries11.7 &47.69 &801  &\bfseries1,204 & {21.3} \\
			\hline
		\end{tabular}}
	\end{center}
	\label{tab:performance_MOT16}
\end{table*}

\begin{table*}[]
\centering
\caption{Evaluations on the MOT17 benchmark from the existing online MOT algorithms.
	$\uparrow$ means the larger the better and $\downarrow$ means the smaller the better.
	The optimal results are shown in \bfseries{bold}.}
\renewcommand\arraystretch{1.4}
\begin{center}
	\setlength{\tabcolsep}{2.1mm}{
		\begin{tabular}{l |c |c | c | c | c | c |c  |c | c | c}
			\hline 
			Methods  
			&\multicolumn{1}{|c|}{MOTA\,$\uparrow$} &\multicolumn{1}{|c|}{MOTP\,$\uparrow$}
			&\multicolumn{1}{|c|}{IDF1\,$\uparrow$}
			&\multicolumn{1}{|c|}{IDR\,$\uparrow$} &\multicolumn{1}{|c|}{FP\,$\downarrow$} 
			&\multicolumn{1}{|c|}{MT\,$\uparrow$}
			&\multicolumn{1}{|c|}{ML\,$\downarrow$}
			&\multicolumn{1}{|c|}{IDS\,$\downarrow$} &\multicolumn{1}{|c}{Frag\,$\downarrow$}
			&\multicolumn{1}{|c}{{Hz}\,$\uparrow$} \\
			\cline{1-11}
			N1T~\citep{baisa2019development}    &42.1 &76.2 &33.9 &47.2 &18,214 &11.9 &42.7 &10,698 &10,864 & {9.9} \\
			SAS~\citep{2020Eliminating}    &44.2 &76.3 &57.2 &49.7 &29,473 &16.1 &44.3 &\bfseries1,529 &\bfseries2,644 & {4.8} \\
			PHD\_DAL~\citep{2019Online}   &44.4 &69.7 &36.2 &49.8 &19,170 &14.9 &39.4 &11,137 &13,900 &{3.4} \\
			HISP~\citep{baisa2021robust}  &45.4 &77.3 &39.9 &50.8 &21,820 &14.8 &39.2 &8,727 &7,147 &{3.2}\\
			GMPHD\_Rd17~\citep{baisa2021occlusion}  &46.8 &75.0 &54.1 &54.3 &38,452 &19.7 &\bfseries33.3 &3,865 &8,097 &\bfseries{30.8}\\
			DASOT17~\citep{chu2020dasot}  &49.5 &75.6 &51.8 &56.2 &33,640 &\bfseries20.4 &34.6 &4,142 &6,852 &{9.1} \\
			GNN~\citep{li2020graph}  &45.5 &76.3 &40.5 &41.8 &25,685 &15.6 &40.6 &4,091 &5,579 &{2.3} \\
			MTP~\citep{kim2021discriminative}    &51.5 &\bfseries77.9 &{54.9} &\bfseries57.2 &29,616 &20.4 &35.5 &2,566 &7,748 &{20.1}\\
			LSST~\citep{feng2019multi}    &52.7 &73.2 &{\bfseries57.9} &57.1 &22,512 &17.9 &36.6 &2,167 &7,443 &{1.8}\\
			Ours    &\bfseries53.5 &77.2 &{50.0} &39.4 &\bfseries10,719 &19.9 &35.8 &2,610  &4,602 &{20.4} \\
			\hline
		\end{tabular}}
	\end{center}
	\label{tab:performance_MOT17}
\end{table*}

\begin{table*}[]
	\centering
	\caption{Evaluations on the MOT20 benchmark from the existing online MOT algorithms.
		$\uparrow$ means the larger the better and $\downarrow$ means the smaller the better.
		The optimal results are shown in \bfseries{bold}.}
	\renewcommand\arraystretch{1.4}
	\begin{center}
		\setlength{\tabcolsep}{2.1mm}{
			\begin{tabular}{l |c |c | c | c | c | c |c  |c | c }
				\hline 
				Methods  
				&\multicolumn{1}{|c|}{MOTA\,$\uparrow$} &\multicolumn{1}{|c|}{MOTP\,$\uparrow$}
				&\multicolumn{1}{|c|}{IDF1\,$\uparrow$}
				&\multicolumn{1}{|c|}{IDR\,$\uparrow$} &\multicolumn{1}{|c|}{MT\,$\uparrow$} &\multicolumn{1}{|c|}{ML\,$\downarrow$}
				&\multicolumn{1}{|c|}{IDS\,$\downarrow$}
				&\multicolumn{1}{|c|}{Frag\,$\downarrow$}
				&\multicolumn{1}{|c}{Hz\,$\uparrow$} \\
				\cline{1-10}
				SORT~\citep{bewley2016simple}   &42.7 &66.9 &45.1  &48.8 &16.7 &26.2 &4,470 & 17,798 &\bfseries57.3 \\
				GMPHD\_Rd20~\citep{baisa2021occlusion} &44.7 &72.0 &43.5 &54.4 &23.6 &22.1 &7,492 &11,153 &25.2 \\
				IOU\_KMM~\citep{urbann2021online}  &46.5 &\bfseries75.2 &49.4 &\bfseries58.5 &\bfseries29.9 &\bfseries19.6 &4,509 &7,557 &30.3 \\
				FlowTracker~\citep{nishimura2021sdof}    &46.7 &68.3 &42.4 &58.0 &27.8 &20.0 &3,532 &5,165 &19.2 \\
				Tracktor++~\citep{bergmann2019tracking}    &52.6 &71.6 &{52.7} &54.3 &29.4 &33.1 &\bfseries1,648 &\bfseries4,374 &1.2 \\
				Ours    &\bfseries52.8 &72.5 &\bfseries53.3 &49.5 &26.3 &21.7 &3,173 &5,718  &15.8 \\
				\hline
		\end{tabular}}
	\end{center}
	\label{tab:performance_MOT20}
\end{table*}

\subsection{Ablation Study} 
\label{sec:ablationstudy} 
Besides, we also conduct some ablation experiments. 
We remove a foundational module each time to prove the effectiveness of each component in the proposed approach as depicted in \Cref{fig:ablation}. 
Each foundational module is indicated below.

\subsubsection{STURE method}
To verify the validity of the designed STURE training approach for object association in MOT task, the designed STURE training method is removed and the original detection and sequence features are utilized to associate the lost target.
In addition, the convolutional operation on current detections is applied and the maximal tracking score on the confidence map is utilized to compute the affinity for object association.
The results of detection-to-sequence object association on the two MOT benchmarks are shown in \Cref{fig:ablation}.

With results from these ablation experiments, it's easy to find that STURE enhances the performance significantly and consistently. 
Specifically, STURE increases the MOTA by 8.1\% on MOT16 and 9.6\% on MOT17{,} respectively, which indicates that temporal representation is crucial for detection-to-sequence feature learning and object association.
The performance results prove that STURE is able to extract spatial-temporal representation effectively from various perspectives and they are mutual complementation.

\begin{figure}[htbp]
	\centering
	
	\subfigure[Without STURE]{
		\begin{minipage}[t]{0.48\linewidth}
			\centering
			\includegraphics[width=1\textwidth]{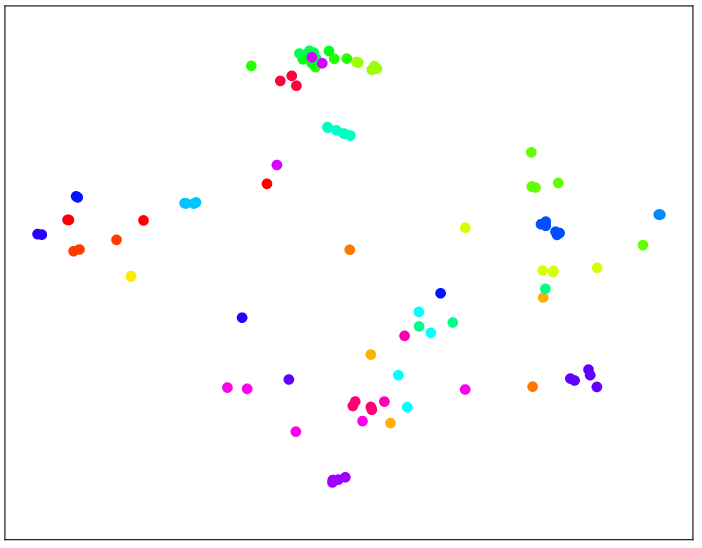}
		\end{minipage}%
	}%
	\subfigure[With STURE]{
		\begin{minipage}[t]{0.48\linewidth}
			\centering
			\includegraphics[width=1\textwidth]{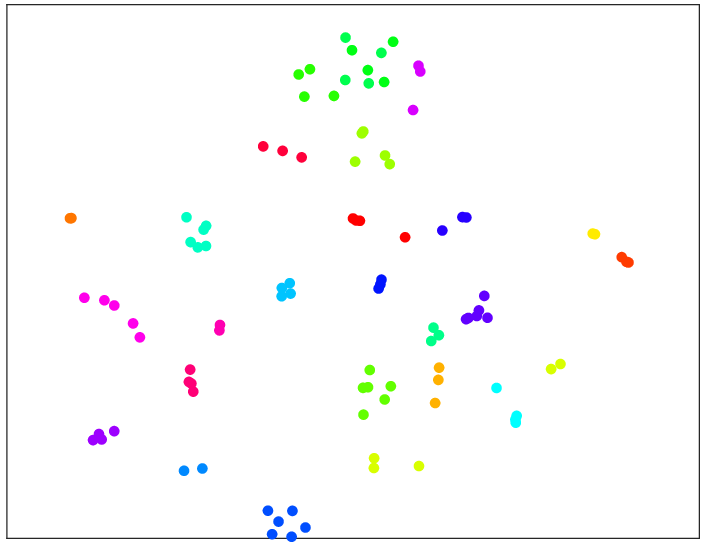}
		\end{minipage}%
	}%
	
	\subfigure[Tracking results without STURE]{
		\begin{minipage}[t]{0.48\linewidth}
			\centering
			\includegraphics[width=1\textwidth]{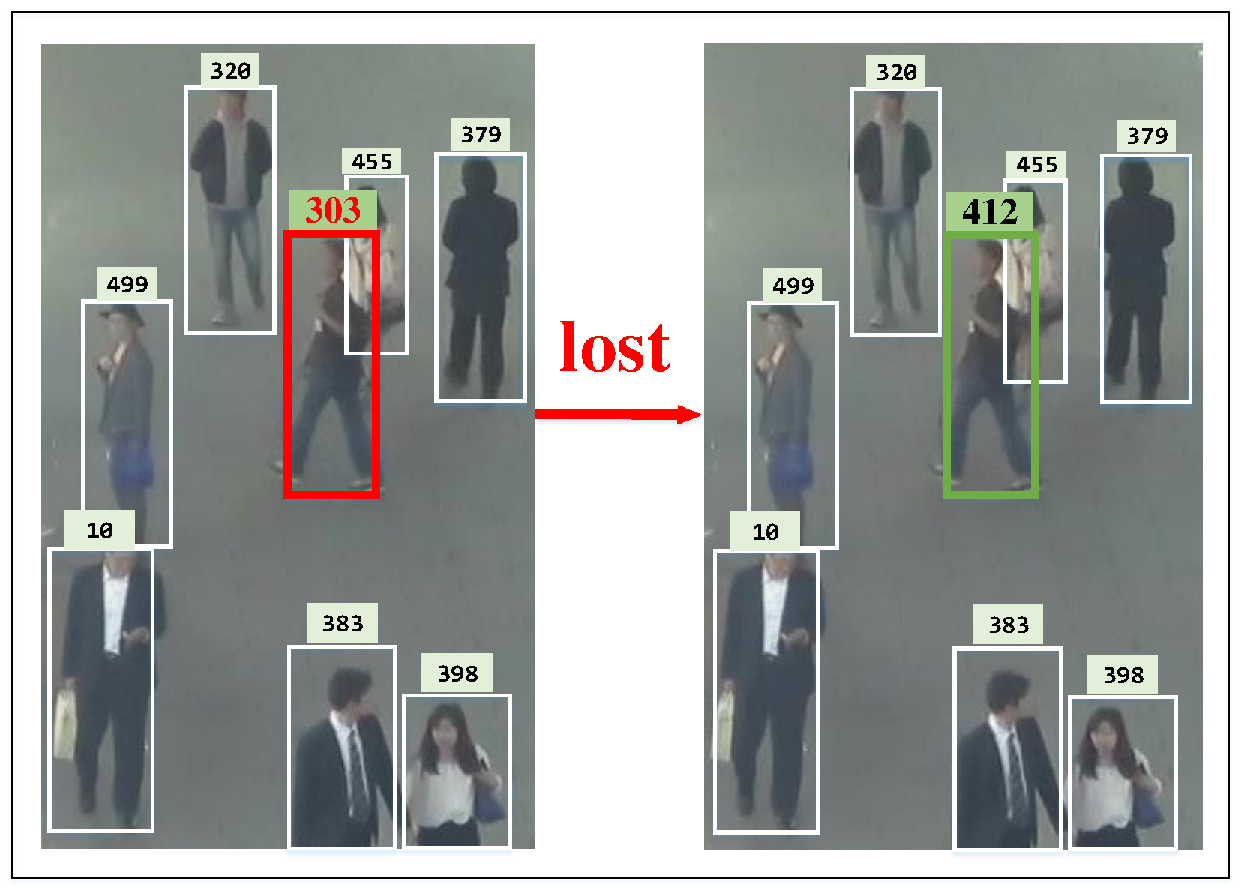}
		\end{minipage}
	}%
	\subfigure[Tracking results with STURE]{
		\begin{minipage}[t]{0.48\linewidth}
			\centering
			\includegraphics[width=1\textwidth]{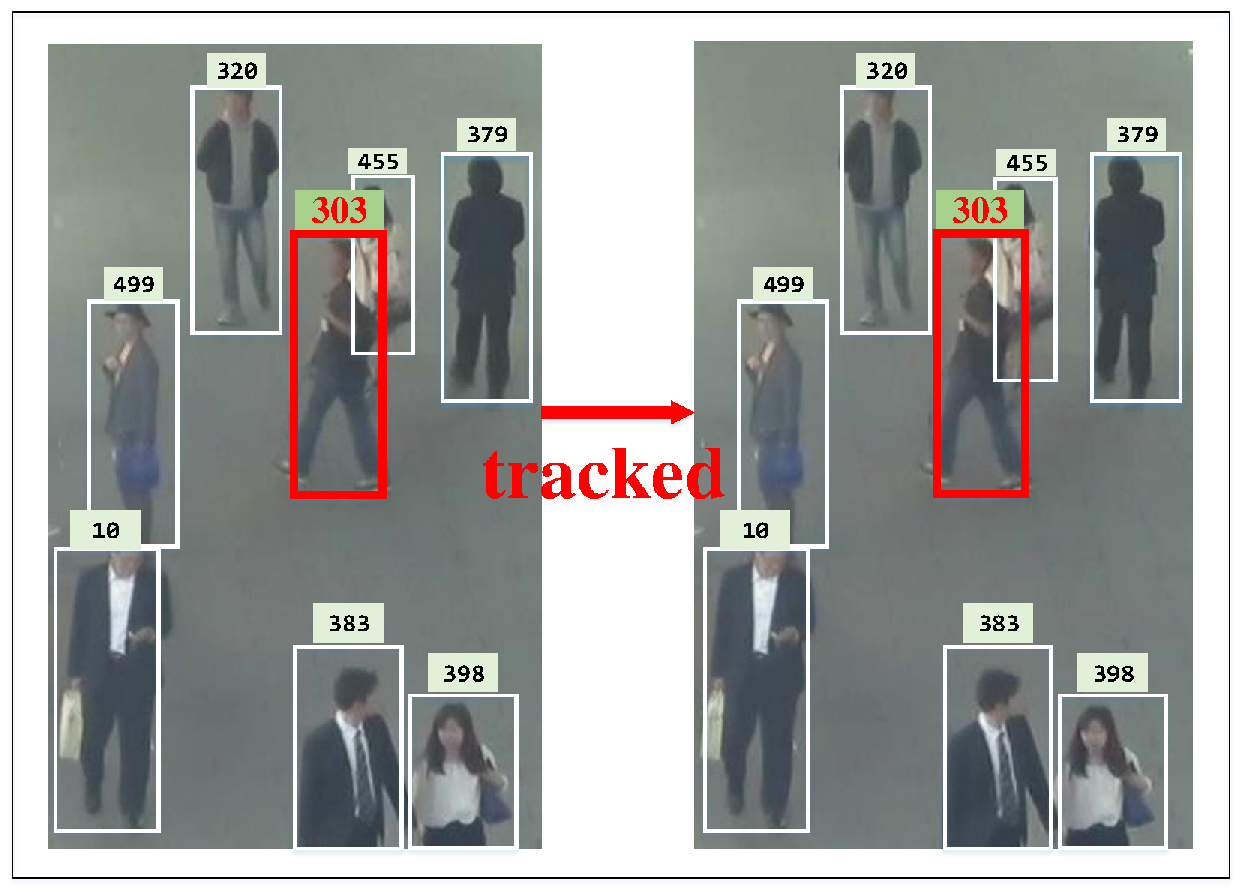}
		\end{minipage}
	}%
	
	\centering
	\caption{The visualisation of learned mutual representation distribution and tracking results without/with STURE using t-SNE~\citep{van2008visualizing} {on} MOT17 dataset.
		Different colors in representation distribution represent different identities. 
	}
	\label{fig:tsne}
\end{figure}

{
We visualize the distribution of the learned mutual representations and corresponding tracking results without/with STURE using t-SNE~\citep{van2008visualizing} as shown in \Cref{fig:tsne}.
{
Every point in \Cref{fig:tsne} (a) and \Cref{fig:tsne} (b) is a 2,048 dimensional feature.}
It is easy to see that the original representations with the same identity are incompact as depicted in \Cref{fig:tsne} (a). 
After STURE, the learned mutual representationss become more consistent as depicted in \Cref{fig:tsne} (b);
and it can improve rubustness to MOT challenges as depicted in \Cref{fig:tsne} (c) and (d). 
Thus, the reinforced representations can improve tracking performance significantly.
}
%
%

\vspace{5pt}
\noindent
\subsubsection{Non-local block}
\begin{figure}[!ht]
	\centering
	\includegraphics[width=1.0\linewidth]{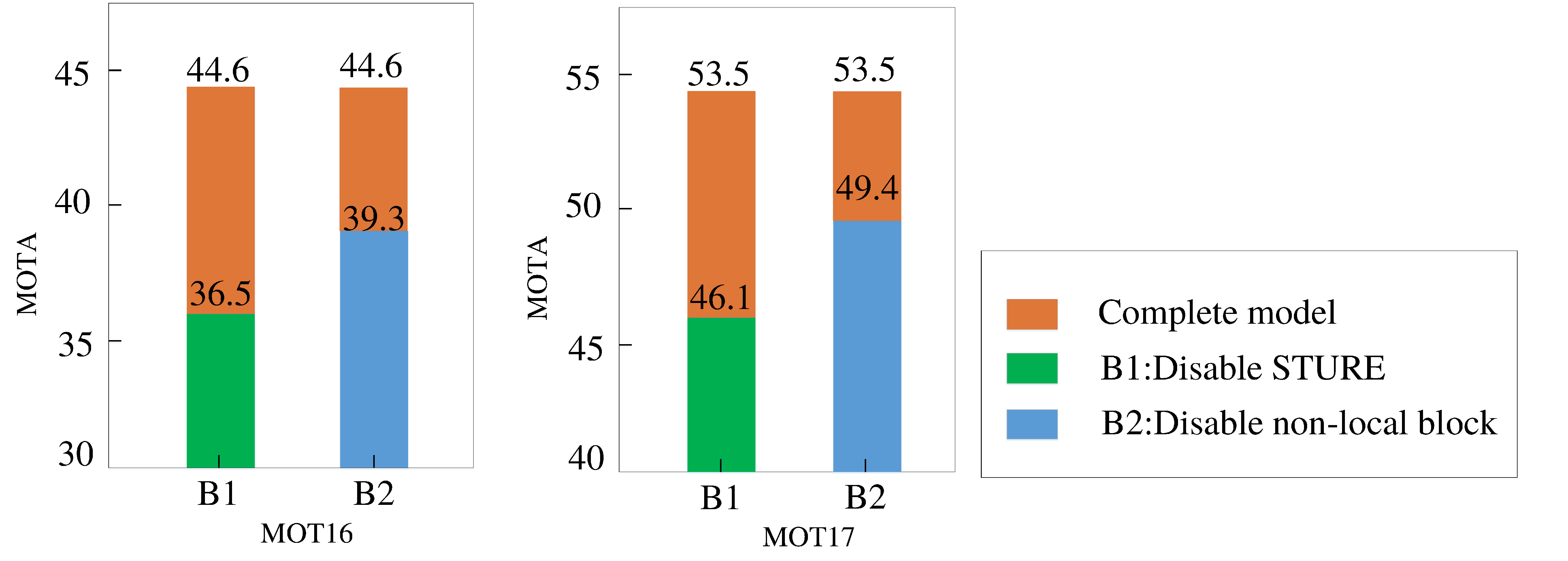}
	\caption{The tracking performance of each ablation method against the designed intact approach (44.6$\%$) on the MOT16 and MOT17 benchmark. 
		It can be seen that each designed module contributes to performance improvement.
		The MOTA decreases sharply by 8.1$\%$ if the tracking score is utilized for MOT, which reflects the merit of STURE. 
		The decline by disabling non-local layers proves that {they} are more valid than the classical convolution architecture.}
	\label{fig:ablation}
\end{figure}

The non-local layers are utilized to extract temporal representations among pedestrian sequence.
To demonstrate the effectiveness of the added non-local layers, we remove them and use the classic CNN structure to learn sequence representations. 
And the sequence frame feature $f_{Sn}$ is substituted with the sequence representation from 3D average pooling.
In \Cref{fig:ablation}, if we delete the non-local layers, the tracking performance in MOT16 and MOT17 still surpasses the ablation status sharply.
However, the tracking performance removing STURE is lower than deleting non-local layers.
Compared with the classical 3D pooling operation, it is believed that non-local layers are able to extract temporal features better,
and they also help the detection learning network learn temporal features more effectively. 
In addition, the proposed STURE is much more important than non-local blocks in improving the tracking performance.

\subsection{Discussion}  
The proposed STURE and robust object association method deal with the trajectory association in online MOT conjunctively. 
The object association is performed between current candidate detections and historical tracklets.

\vspace{5pt}
\noindent
\subsubsection{Sequence Size}
\begin{figure}[!ht]
	\centering
	\includegraphics[width=1.0\linewidth]{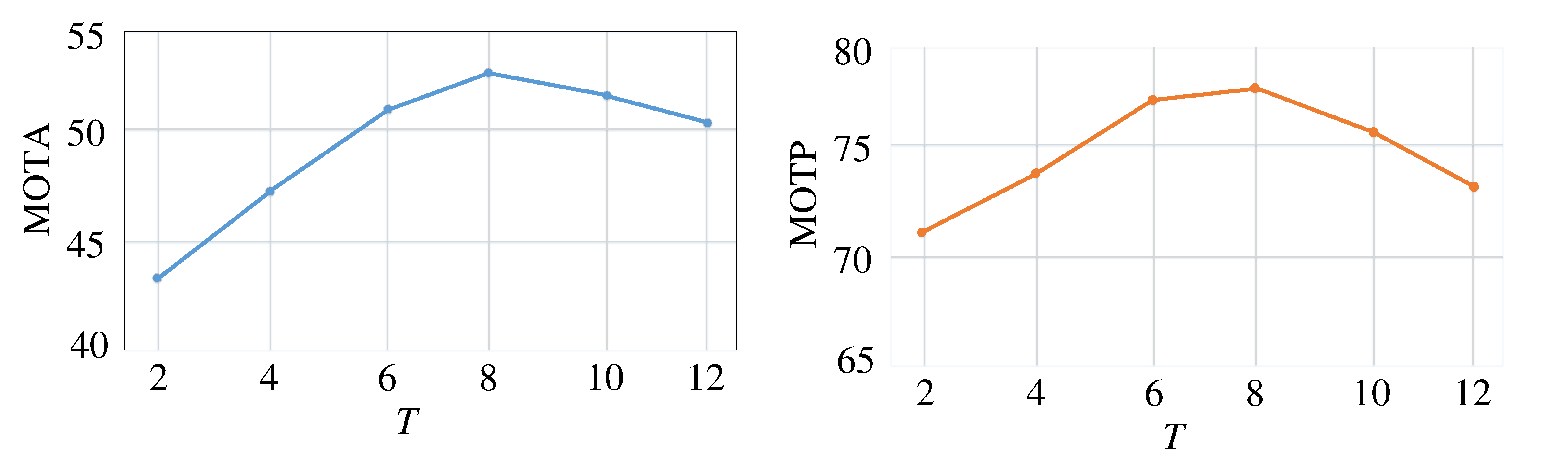}
	\caption{The MOTA and MOTP results with different $T$ on the MOT16 dataset.}
	\label{fig:T}
\end{figure}

The number of pedestrian sequence images is a crucial parameter for tracking performance in the designed architecture.
The performance experiment with different values of $T$ is showed in Figure~\ref{fig:T} .
It is easy to find that, when $T$ is set to $8$, the optimal MOTA and MOTP are obtained simultaneously.

\vspace{5pt}
\noindent
\subsubsection{Tracking hyper-parameters}
In addition, several experiments are conducted to indicate the effect of various thresholds, such as the trajectory launch, trajectory termination, appearance similarity, tracking score and IoU.
The tracking parameters are tested on various parametric settings. 
The MOTA varies drastically with the settings of $\tau_s$ and $\tau_a$. 
By this means, $ \tau _s $ and $ \tau _a $ are selected with grid search based on the trained association network. 
When $ \tau _s=0.2 $ and $ \tau _a=0.8 $, the maximized MOTA is gained. 
Therefore, the tracking parameters are selected as shown before.

\begin{figure*}[!ht]
	\centering
	\includegraphics[width=0.99\linewidth]{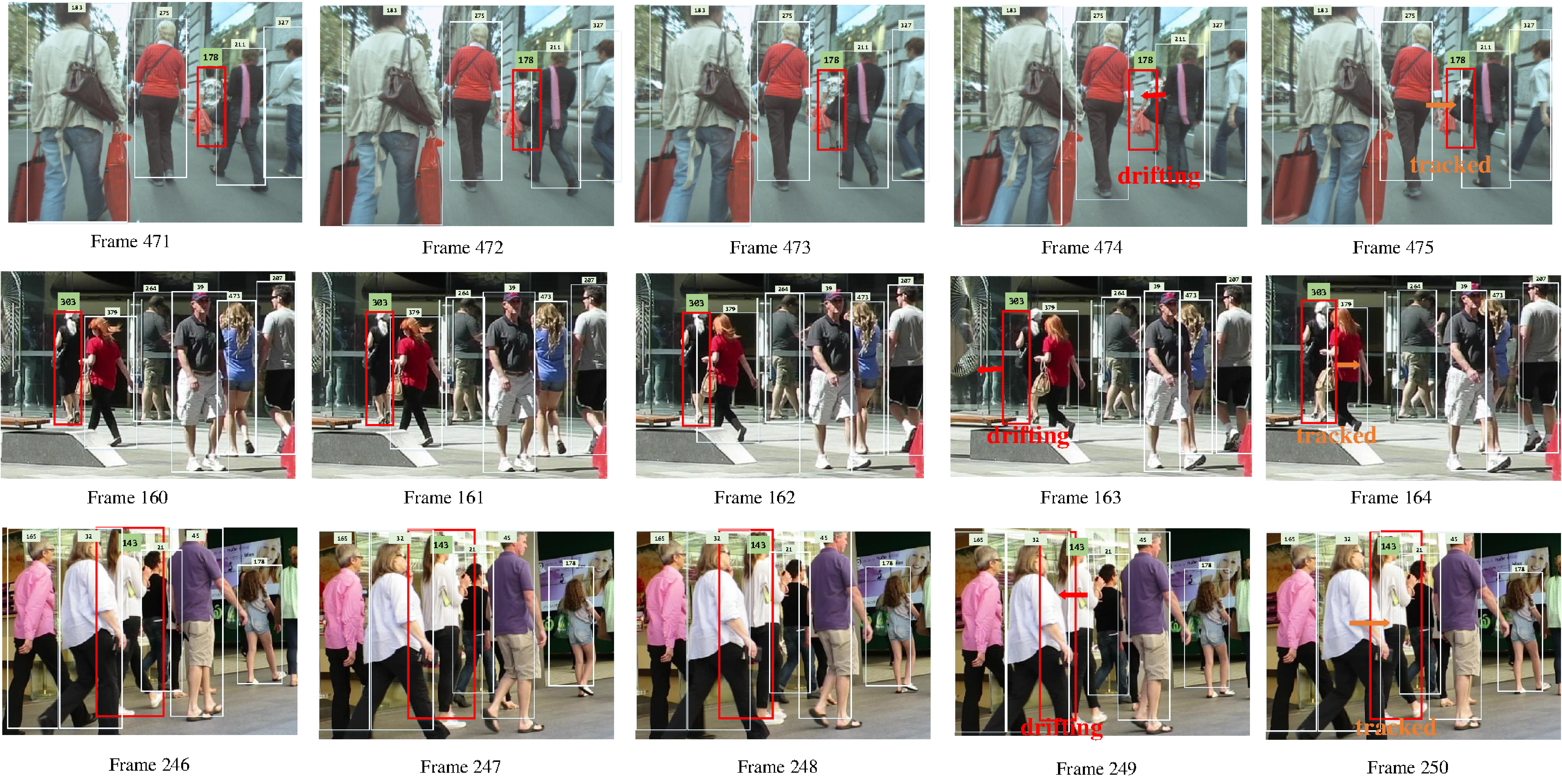}
	\caption{Example sequences of "MOT17-05" on the first row (street scene from a moving platform), {"MOT17-08" in the second row (a crowded pedestrian street, stationary camera)} and "MOT17-09" in the third row (a crowded pedestrian environment filmed from a lower viewpoint) show that the proposed algorithm can associate objects when the single object tracking process turns into drifting status in the penultimate column.}
	\label{fig:tracking_result}
\end{figure*}

The tracking {demonstrations are} shown in \Cref{fig:tracking_result}. 
In general, the single object tracker is easy to drift when the object moves fast in camera or is affected by others. 
In this work, a robust association approach is utilized to deal with the drifting. 
Besides, the single object tracker is able to solve the problems of occlusion well.

\section{Conclusion}
In this study, a novel STURE method is proposed for robust object association in online MOT task. 
The STURE schema can learn temporal representation in the sequence learning network and the detection learning network mutually.
Using the mutual learning, the detection learning network can learn temporal features and become more robust, and the feature difference between current detections and historical sequences will be relieved as well.
And it is useful to associate current detection {results} and historical sequences to deal with imperfect {detections}.
Compared with the existing online MOT methods, the proposed method can get better tracking results on the MOT benchmarks,
and extensive experiments prove the effectiveness of the designed approach.

\section*{Acknowledgments}
This work was partially supported by National Key Research and Development Program of China (No.2018YFB1308604), National Natural Science Foundation of China (No.61976086,62106071), Hunan Innovation Technology Investment Project (No.2019GK5061), and Special Project of Foshan Science and Technology Innovation Team (No. FS0AA-KJ919-4402-0069).

%

%

\bibliographystyle{model2-names}
\bibliography{ycviu-template-with-authorship}

\begin{thebibliography}{65}
\expandafter\ifx\csname natexlab\endcsname\relax\def\natexlab#1{#1}\fi
\providecommand{\url}[1]{\texttt{#1}}
\providecommand{\href}[2]{#2}
\providecommand{\path}[1]{#1}
\providecommand{\DOIprefix}{doi:}
\providecommand{\ArXivprefix}{arXiv:}
\providecommand{\URLprefix}{URL: }
\providecommand{\Pubmedprefix}{pmid:}
\providecommand{\doi}[1]{\href{http://dx.doi.org/#1}{\path{#1}}}
\providecommand{\Pubmed}[1]{\href{pmid:#1}{\path{#1}}}
\providecommand{\bibinfo}[2]{#2}
\ifx\xfnm\relax \def\xfnm[#1]{\unskip,\space#1}\fi
\bibitem[{Andreas~Geiger and Urtasun(2012)}]{geiger2012are}
\bibinfo{author}{Andreas~Geiger, P.L.}, \bibinfo{author}{Urtasun, R.},
  \bibinfo{year}{2012}.
\newblock \bibinfo{title}{Are we ready for autonomous driving? the kitti vision
  benchmark suite}.
\newblock \bibinfo{journal}{Computer Vision and Pattern Recognition} ,
  \bibinfo{pages}{3354--3361}.
\bibitem[{Bae and Yoon(2014)}]{bae2014robust}
\bibinfo{author}{Bae, S.H.}, \bibinfo{author}{Yoon, K.J.},
  \bibinfo{year}{2014}.
\newblock \bibinfo{title}{Robust online multi-object tracking based on tracklet
  confidence and online discriminative appearance learning}.
\newblock \bibinfo{journal}{CVPR} , \bibinfo{pages}{1218--1225}.
\bibitem[{Bae and Yoon(2018)}]{RN601}
\bibinfo{author}{Bae, S.H.}, \bibinfo{author}{Yoon, K.J.},
  \bibinfo{year}{2018}.
\newblock \bibinfo{title}{Confidence-based data association and discriminative
  deep appearance learning for robust online multi-object tracking}.
\newblock \bibinfo{journal}{IEEE Transactions on Pattern Analysis and Machine
  Intelligence} , \bibinfo{pages}{1--1}.
\bibitem[{Baisa and Wallace(2019)}]{baisa2019development}
\bibinfo{author}{Baisa, L.N.}, \bibinfo{author}{Wallace, M.A.},
  \bibinfo{year}{2019}.
\newblock \bibinfo{title}{Development of a n-type gm-phd filter for multiple
  target, multiple type visual tracking}.
\newblock \bibinfo{journal}{Journal of Visual Communication and Image
  Representation} , \bibinfo{pages}{257--271}.
\bibitem[{Baisa(2019)}]{2019Online}
\bibinfo{author}{Baisa, N.L.}, \bibinfo{year}{2019}.
\newblock \bibinfo{title}{Online multi-object visual tracking using a gm-phd
  filter with deep appearance learning}, in: \bibinfo{booktitle}{22nd
  International Conference on Information Fusion}.
\bibitem[{Baisa(2021a)}]{baisa2021occlusion}
\bibinfo{author}{Baisa, N.L.}, \bibinfo{year}{2021}a.
\newblock \bibinfo{title}{Occlusion-robust online multi-object visual tracking
  using a gm-phd filter with cnn-based re-identification}.
\newblock \bibinfo{journal}{Journal of Visual Communication and Image
  Representation} \bibinfo{volume}{80}, \bibinfo{pages}{103279}.
\bibitem[{Baisa(2021b)}]{baisa2021robust}
\bibinfo{author}{Baisa, N.L.}, \bibinfo{year}{2021}b.
\newblock \bibinfo{title}{Robust online multi-target visual tracking using a
  hisp filter with discriminative deep appearance learning}.
\newblock \bibinfo{journal}{Journal of Visual Communication and Image
  Representation} \bibinfo{volume}{77}, \bibinfo{pages}{102952}.
\bibitem[{Bergmann et~al.(2019)Bergmann, Meinhardt and
  Leal-Taixe}]{bergmann2019tracking}
\bibinfo{author}{Bergmann, P.}, \bibinfo{author}{Meinhardt, T.},
  \bibinfo{author}{Leal-Taixe, L.}, \bibinfo{year}{2019}.
\newblock \bibinfo{title}{Tracking without bells and whistles}, in:
  \bibinfo{booktitle}{Proceedings of the IEEE/CVF International Conference on
  Computer Vision}, pp. \bibinfo{pages}{941--951}.
\bibitem[{Bernardin and Stiefelhagen(2008)}]{RN475}
\bibinfo{author}{Bernardin, K.}, \bibinfo{author}{Stiefelhagen, R.},
  \bibinfo{year}{2008}.
\newblock \bibinfo{title}{Evaluating multiple object tracking performance: the
  clear mot metrics}.
\newblock \bibinfo{journal}{Journal on Image and Video Processing}
  \bibinfo{volume}{2008}, \bibinfo{pages}{1}.
\bibitem[{Bewley et~al.(2016)Bewley, Ge, Ott, Ramos and
  Upcroft}]{bewley2016simple}
\bibinfo{author}{Bewley, A.}, \bibinfo{author}{Ge, Z.}, \bibinfo{author}{Ott,
  L.}, \bibinfo{author}{Ramos, F.}, \bibinfo{author}{Upcroft, B.},
  \bibinfo{year}{2016}.
\newblock \bibinfo{title}{Simple online and realtime tracking}, in:
  \bibinfo{booktitle}{2016 IEEE international conference on image processing
  (ICIP)}, \bibinfo{organization}{IEEE}. pp. \bibinfo{pages}{3464--3468}.
\bibitem[{Chen et~al.(2018)Chen, Wang and Zhang}]{RN982}
\bibinfo{author}{Chen, Y.}, \bibinfo{author}{Wang, N.}, \bibinfo{author}{Zhang,
  Z.}, \bibinfo{year}{2018}.
\newblock \bibinfo{title}{Darkrank: Accelerating deep metric learning via cross
  sample similarities transfer}.
\newblock \bibinfo{journal}{AAAI} , \bibinfo{pages}{2852--2859}.
\bibitem[{Chen et~al.(2019)Chen, Yuan, Li, Wu, Nouioua and Xie}]{RN724}
\bibinfo{author}{Chen, Y.}, \bibinfo{author}{Yuan, J.}, \bibinfo{author}{Li,
  Z.}, \bibinfo{author}{Wu, Y.}, \bibinfo{author}{Nouioua, M.},
  \bibinfo{author}{Xie, G.}, \bibinfo{year}{2019}.
\newblock \bibinfo{title}{Person re-identification based on re-ranking with
  expanded k-reciprocal nearest neighbors}.
\newblock \bibinfo{journal}{Journal of Visual Communication and Image
  Representation} \bibinfo{volume}{58}, \bibinfo{pages}{486--494}.
\bibitem[{Cheng et~al.(2016)Cheng, Gong, Zhou, Wang and
  Zheng}]{cheng2016person}
\bibinfo{author}{Cheng, D.}, \bibinfo{author}{Gong, Y.}, \bibinfo{author}{Zhou,
  S.}, \bibinfo{author}{Wang, J.}, \bibinfo{author}{Zheng, N.},
  \bibinfo{year}{2016}.
\newblock \bibinfo{title}{Person re-identification by multi-channel parts-based
  cnn with improved triplet loss function}.
\newblock \bibinfo{journal}{CVPR} , \bibinfo{pages}{1335--1344}.
\bibitem[{Chu and Ling(2019)}]{RN542}
\bibinfo{author}{Chu, P.}, \bibinfo{author}{Ling, H.}, \bibinfo{year}{2019}.
\newblock \bibinfo{title}{Famnet: Joint learning of feature, affinity and
  multi-dimensional assignment for online multiple object tracking}.
\newblock \bibinfo{journal}{International Conference on Computer Vision} ,
  \bibinfo{pages}{6172--6181}.
\bibitem[{Chu et~al.(2017)Chu, Ouyang, Li, Wang, Liu and Yu}]{chu2017online}
\bibinfo{author}{Chu, Q.}, \bibinfo{author}{Ouyang, W.}, \bibinfo{author}{Li,
  H.}, \bibinfo{author}{Wang, X.}, \bibinfo{author}{Liu, B.},
  \bibinfo{author}{Yu, N.}, \bibinfo{year}{2017}.
\newblock \bibinfo{title}{Online multi-object tracking using cnn-based single
  object tracker with spatial-temporal attention mechanism}.
\newblock \bibinfo{journal}{ICCV} , \bibinfo{pages}{4846--4855}.
\bibitem[{Chu et~al.(2020)Chu, Ouyang, Liu, Zhu and Yu}]{chu2020dasot}
\bibinfo{author}{Chu, Q.}, \bibinfo{author}{Ouyang, W.}, \bibinfo{author}{Liu,
  B.}, \bibinfo{author}{Zhu, F.}, \bibinfo{author}{Yu, N.},
  \bibinfo{year}{2020}.
\newblock \bibinfo{title}{Dasot: A unified framework integrating data
  association and single object tracking for online multi-object tracking}.
\newblock \bibinfo{journal}{AAAI} , \bibinfo{pages}{10672--10679}.
\bibitem[{Chung et~al.(2017)Chung, Tahboub and Delp}]{chung2017a}
\bibinfo{author}{Chung, D.}, \bibinfo{author}{Tahboub, K.},
  \bibinfo{author}{Delp, J.E.}, \bibinfo{year}{2017}.
\newblock \bibinfo{title}{A two stream siamese convolutional neural network for
  person re-identification}.
\newblock \bibinfo{journal}{ICCV} , \bibinfo{pages}{1992--2000}.
\bibitem[{Danelljan et~al.(2017)Danelljan, Bhat, Khan and
  Felsberg}]{danelljan2017eco}
\bibinfo{author}{Danelljan, M.}, \bibinfo{author}{Bhat, G.},
  \bibinfo{author}{Khan, S.F.}, \bibinfo{author}{Felsberg, M.},
  \bibinfo{year}{2017}.
\newblock \bibinfo{title}{Eco: Efficient convolution operators for tracking}.
\newblock \bibinfo{journal}{CVPR} .
\bibitem[{Dehghan et~al.(2015)Dehghan, Modiri~Assari and
  Shah}]{dehghan2015gmmcp}
\bibinfo{author}{Dehghan, A.}, \bibinfo{author}{Modiri~Assari, S.},
  \bibinfo{author}{Shah, M.}, \bibinfo{year}{2015}.
\newblock \bibinfo{title}{Gmmcp tracker: Globally optimal generalized maximum
  multi clique problem for multiple object tracking}.
\newblock \bibinfo{journal}{CVPR} .
\bibitem[{Dendorfer et~al.(2019)Dendorfer, Rezatofighi, Milan, Shi, Cremers,
  Reid, Roth, Schindler and Leal-Taixe}]{2019CVPR19}
\bibinfo{author}{Dendorfer, P.}, \bibinfo{author}{Rezatofighi, H.},
  \bibinfo{author}{Milan, A.}, \bibinfo{author}{Shi, J.},
  \bibinfo{author}{Cremers, D.}, \bibinfo{author}{Reid, I.},
  \bibinfo{author}{Roth, S.}, \bibinfo{author}{Schindler, K.},
  \bibinfo{author}{Leal-Taixe, L.}, \bibinfo{year}{2019}.
\newblock \bibinfo{title}{Cvpr19 tracking and detection challenge: How crowded
  can it get?} .
\bibitem[{Felzenszwalb et~al.(2010)Felzenszwalb, Girshick, Mcallester and
  Ramanan}]{RN584}
\bibinfo{author}{Felzenszwalb, P.F.}, \bibinfo{author}{Girshick, R.B.},
  \bibinfo{author}{Mcallester, D.}, \bibinfo{author}{Ramanan, D.},
  \bibinfo{year}{2010}.
\newblock \bibinfo{title}{Object detection with discriminatively trained
  part-based models}.
\newblock \bibinfo{journal}{IEEE Transactions on Pattern Analysis \& Machine
  Intelligence} \bibinfo{volume}{32}, \bibinfo{pages}{1627--1645}.
\bibitem[{Feng et~al.(2019)Feng, Hu, Wu, Yan and Ouyang}]{feng2019multi}
\bibinfo{author}{Feng, W.}, \bibinfo{author}{Hu, Z.}, \bibinfo{author}{Wu, W.},
  \bibinfo{author}{Yan, J.}, \bibinfo{author}{Ouyang, W.},
  \bibinfo{year}{2019}.
\newblock \bibinfo{title}{Multi-object tracking with multiple cues and
  switcher-aware classification}.
\newblock \bibinfo{journal}{arXiv preprint arXiv:1901.06129} .
\bibitem[{Gu et~al.(2019)Gu, Ma, Chang, Shan and Chen}]{RN927}
\bibinfo{author}{Gu, X.}, \bibinfo{author}{Ma, B.}, \bibinfo{author}{Chang,
  H.}, \bibinfo{author}{Shan, S.}, \bibinfo{author}{Chen, X.},
  \bibinfo{year}{2019}.
\newblock \bibinfo{title}{Temporal knowledge propagation for image-to-video
  person re-identification}.
\newblock \bibinfo{journal}{ICCV} , \bibinfo{pages}{9646--9655}.
\bibitem[{Hinton et~al.(2015)Hinton, Vinyals and Dean}]{hinton2015distilling}
\bibinfo{author}{Hinton, E.G.}, \bibinfo{author}{Vinyals, O.},
  \bibinfo{author}{Dean, J.}, \bibinfo{year}{2015}.
\newblock \bibinfo{title}{Distilling the knowledge in a neural network}.
\newblock \bibinfo{journal}{CoRR} .
\bibitem[{Junbo et~al.(2020)Junbo, Wenguan, Qinghao, Ruigang and
  Jianbing}]{junbo2020a}
\bibinfo{author}{Junbo, Y.}, \bibinfo{author}{Wenguan, W.},
  \bibinfo{author}{Qinghao, M.}, \bibinfo{author}{Ruigang, Y.},
  \bibinfo{author}{Jianbing, S.}, \bibinfo{year}{2020}.
\newblock \bibinfo{title}{A unified object motion and affinity model for online
  multi-object tracking}.
\newblock \bibinfo{journal}{CVPR} , \bibinfo{pages}{6767--6776}.
\bibitem[{Kaiming~He and Sun(2016)}]{he2016deep}
\bibinfo{author}{Kaiming~He, Xiangyu~Zhang, S.R.}, \bibinfo{author}{Sun, J.},
  \bibinfo{year}{2016}.
\newblock \bibinfo{title}{Deep residual learning for image recognition}.
\newblock \bibinfo{journal}{CVPR} .
\bibitem[{Kim et~al.(2021)Kim, Fuxin, Alotaibi and
  Rehg}]{kim2021discriminative}
\bibinfo{author}{Kim, C.}, \bibinfo{author}{Fuxin, L.},
  \bibinfo{author}{Alotaibi, M.}, \bibinfo{author}{Rehg, J.M.},
  \bibinfo{year}{2021}.
\newblock \bibinfo{title}{Discriminative appearance modeling with multi-track
  pooling for real-time multi-object tracking}, in:
  \bibinfo{booktitle}{Proceedings of the IEEE/CVF Conference on Computer Vision
  and Pattern Recognition}, pp. \bibinfo{pages}{9553--9562}.
\bibitem[{Kingma and Ba(2014)}]{RN586}
\bibinfo{author}{Kingma, D.}, \bibinfo{author}{Ba, J.}, \bibinfo{year}{2014}.
\newblock \bibinfo{title}{Adam: A method for stochastic optimization} .
\bibitem[{Leal-Taixé et~al.(2016)Leal-Taixé, Canton-Ferrer and
  Schindler}]{leal-taixe2016learning}
\bibinfo{author}{Leal-Taixé, L.}, \bibinfo{author}{Canton-Ferrer, C.},
  \bibinfo{author}{Schindler, K.}, \bibinfo{year}{2016}.
\newblock \bibinfo{title}{Learning by tracking: Siamese cnn for robust target
  association}.
\newblock \bibinfo{journal}{IEEE Computer Society Conference on Computer Vision
  and Pattern Recognition Workshops} , \bibinfo{pages}{418--425}.
\bibitem[{Levin(1998)}]{RN987}
\bibinfo{author}{Levin, D.}, \bibinfo{year}{1998}.
\newblock \bibinfo{title}{The approximation power of moving least-squares}.
\newblock \bibinfo{journal}{Math. Comput.} , \bibinfo{pages}{1517--1531}.
\bibitem[{Li et~al.(2019)Li, Peng, Nai, Li and Li}]{RN1214}
\bibinfo{author}{Li, G.}, \bibinfo{author}{Peng, M.}, \bibinfo{author}{Nai,
  K.}, \bibinfo{author}{Li, Z.}, \bibinfo{author}{Li, K.},
  \bibinfo{year}{2019}.
\newblock \bibinfo{title}{Multi-view correlation tracking with adaptive
  memory-improved update model}.
\newblock \bibinfo{journal}{Neural Computing and Applications} ,
  \bibinfo{pages}{1--17}.
\bibitem[{Li et~al.(2020a)Li, Gao and Jiang}]{li2020graph}
\bibinfo{author}{Li, J.}, \bibinfo{author}{Gao, X.}, \bibinfo{author}{Jiang,
  T.}, \bibinfo{year}{2020}a.
\newblock \bibinfo{title}{Graph networks for multiple object tracking}.
\newblock \bibinfo{journal}{WACV} , \bibinfo{pages}{708--717}.
\bibitem[{Li et~al.(2021)Li, Lv, Chen and Yuan}]{2021Person}
\bibinfo{author}{Li, Z.}, \bibinfo{author}{Lv, J.}, \bibinfo{author}{Chen, Y.},
  \bibinfo{author}{Yuan, J.}, \bibinfo{year}{2021}.
\newblock \bibinfo{title}{Person re-identification with part prediction
  alignment}.
\newblock \bibinfo{journal}{Computer Vision and Image Understanding}
  \bibinfo{volume}{205}, \bibinfo{pages}{103172}.
\bibitem[{Li et~al.(2020b)Li, Nai, Li and Jiang}]{RN1215}
\bibinfo{author}{Li, Z.}, \bibinfo{author}{Nai, K.}, \bibinfo{author}{Li, G.},
  \bibinfo{author}{Jiang, S.}, \bibinfo{year}{2020}b.
\newblock \bibinfo{title}{Learning a dynamic feature fusion tracker for object
  tracking}.
\newblock \bibinfo{journal}{IEEE Transactions on Intelligent Transportation
  Systems} .
\bibitem[{Li et~al.(2016)Li, Wang, Nai, Shen and Zeng}]{li2016robust}
\bibinfo{author}{Li, Z.}, \bibinfo{author}{Wang, D.}, \bibinfo{author}{Nai,
  K.}, \bibinfo{author}{Shen, T.}, \bibinfo{author}{Zeng, Y.},
  \bibinfo{year}{2016}.
\newblock \bibinfo{title}{Robust object tracking via weight-based local sparse
  appearance model}.
\newblock \bibinfo{journal}{ICNC-FSKD} , \bibinfo{pages}{560--565}.
\bibitem[{Liu et~al.(2019)Liu, Liu, Wu, Li and Yu}]{2019Real}
\bibinfo{author}{Liu, Q.}, \bibinfo{author}{Liu, B.}, \bibinfo{author}{Wu, Y.},
  \bibinfo{author}{Li, W.}, \bibinfo{author}{Yu, N.}, \bibinfo{year}{2019}.
\newblock \bibinfo{title}{Real-time online multi-object tracking in compressed
  domain}.
\newblock \bibinfo{journal}{IEEE Access} .
\bibitem[{Lv et~al.(2020)Lv, Li, Nai, Chen and Yuan}]{RN994}
\bibinfo{author}{Lv, J.}, \bibinfo{author}{Li, Z.}, \bibinfo{author}{Nai, K.},
  \bibinfo{author}{Chen, Y.}, \bibinfo{author}{Yuan, J.}, \bibinfo{year}{2020}.
\newblock \bibinfo{title}{Person re-identification with expanded neighborhoods
  distance re-ranking}.
\newblock \bibinfo{journal}{Image and Vision Computing} \bibinfo{volume}{95},
  \bibinfo{pages}{103875}.
\bibitem[{Van~der Maaten and Hinton(2008)}]{van2008visualizing}
\bibinfo{author}{Van~der Maaten, L.}, \bibinfo{author}{Hinton, G.},
  \bibinfo{year}{2008}.
\newblock \bibinfo{title}{Visualizing data using t-sne.}
\newblock \bibinfo{journal}{Journal of machine learning research}
  \bibinfo{volume}{9}.
\bibitem[{Maksai and Fua(2020)}]{2020Eliminating}
\bibinfo{author}{Maksai, A.}, \bibinfo{author}{Fua, P.}, \bibinfo{year}{2020}.
\newblock \bibinfo{title}{Eliminating exposure bias and metric mismatch in
  multiple object tracking}, in: \bibinfo{booktitle}{2019 IEEE/CVF Conference
  on Computer Vision and Pattern Recognition (CVPR)}.
\bibitem[{McLaughlin et~al.(2016)McLaughlin, Rincón and
  Miller}]{mclaughlin2016recurrent}
\bibinfo{author}{McLaughlin, N.}, \bibinfo{author}{Rincón, M.d.J.},
  \bibinfo{author}{Miller, C.P.}, \bibinfo{year}{2016}.
\newblock \bibinfo{title}{Recurrent convolutional network for video-based
  person re-identification}.
\newblock \bibinfo{journal}{CVPR} , \bibinfo{pages}{1325--1334}.
\bibitem[{Milan et~al.(2016)Milan, Leal-Taixe, Reid, Roth and
  Schindler}]{RN583}
\bibinfo{author}{Milan, A.}, \bibinfo{author}{Leal-Taixe, L.},
  \bibinfo{author}{Reid, I.}, \bibinfo{author}{Roth, S.},
  \bibinfo{author}{Schindler, K.}, \bibinfo{year}{2016}.
\newblock \bibinfo{title}{Mot16: A benchmark for multi-object tracking} .
\bibitem[{Nai et~al.(2018)Nai, Li, Li and Wang}]{RN717}
\bibinfo{author}{Nai, K.}, \bibinfo{author}{Li, Z.}, \bibinfo{author}{Li, G.},
  \bibinfo{author}{Wang, S.}, \bibinfo{year}{2018}.
\newblock \bibinfo{title}{Robust object tracking via local sparse appearance
  model}.
\newblock \bibinfo{journal}{IEEE Transactions on Image Processing}
  \bibinfo{volume}{27}, \bibinfo{pages}{4958--4970}.
\bibitem[{Nai et~al.(2019)Nai, Xiao, Li, Jiang and Gu}]{RN1212}
\bibinfo{author}{Nai, K.}, \bibinfo{author}{Xiao, D.}, \bibinfo{author}{Li,
  Z.}, \bibinfo{author}{Jiang, S.}, \bibinfo{author}{Gu, Y.},
  \bibinfo{year}{2019}.
\newblock \bibinfo{title}{Multi-pattern correlation tracking}.
\newblock \bibinfo{journal}{Knowledge-Based Systems} \bibinfo{volume}{181},
  \bibinfo{pages}{104789}.
\bibitem[{Nishimura et~al.(2021)Nishimura, Komorita, Kawanishi and
  Murase}]{nishimura2021sdof}
\bibinfo{author}{Nishimura, H.}, \bibinfo{author}{Komorita, S.},
  \bibinfo{author}{Kawanishi, Y.}, \bibinfo{author}{Murase, H.},
  \bibinfo{year}{2021}.
\newblock \bibinfo{title}{Sdof-tracker: Fast and accurate multiple human
  tracking by skipped-detection and optical-flow}.
\newblock \bibinfo{journal}{arXiv preprint arXiv:2106.14259} .
\bibitem[{Paszke et~al.(2019)Paszke, Gross, Massa, Lerer, Bradbury, Chanan,
  Killeen, Lin, Gimelshein, Antiga, Desmaison, Köpf, Yang, DeVito, Raison,
  Tejani, Chilamkurthy, Steiner, Fang, Bai and Chintala}]{paszke2019pytorch}
\bibinfo{author}{Paszke, A.}, \bibinfo{author}{Gross, S.},
  \bibinfo{author}{Massa, F.}, \bibinfo{author}{Lerer, A.},
  \bibinfo{author}{Bradbury, J.}, \bibinfo{author}{Chanan, G.},
  \bibinfo{author}{Killeen, T.}, \bibinfo{author}{Lin, Z.},
  \bibinfo{author}{Gimelshein, N.}, \bibinfo{author}{Antiga, L.},
  \bibinfo{author}{Desmaison, A.}, \bibinfo{author}{Köpf, A.},
  \bibinfo{author}{Yang, E.}, \bibinfo{author}{DeVito, Z.},
  \bibinfo{author}{Raison, M.}, \bibinfo{author}{Tejani, A.},
  \bibinfo{author}{Chilamkurthy, S.}, \bibinfo{author}{Steiner, B.},
  \bibinfo{author}{Fang, L.}, \bibinfo{author}{Bai, J.},
  \bibinfo{author}{Chintala, S.}, \bibinfo{year}{2019}.
\newblock \bibinfo{title}{Pytorch - an imperative style, high-performance deep
  learning library}.
\newblock \bibinfo{journal}{NeurIPS} , \bibinfo{pages}{8024--8035}.
\bibitem[{Ren et~al.(2017)Ren, He, Girshick and Sun}]{ren2017faster}
\bibinfo{author}{Ren, S.}, \bibinfo{author}{He, K.}, \bibinfo{author}{Girshick,
  B.R.}, \bibinfo{author}{Sun, J.}, \bibinfo{year}{2017}.
\newblock \bibinfo{title}{Faster r-cnn: Towards real-time object detection with
  region proposal networks}.
\newblock \bibinfo{journal}{IEEE Trans. Pattern Anal. Mach. Intell.} ,
  \bibinfo{pages}{1137--1149}.
\bibitem[{Ristani et~al.(2016)Ristani, Solera, Zou, Cucchiara and
  Tomasi}]{ristani2016performance}
\bibinfo{author}{Ristani, E.}, \bibinfo{author}{Solera, F.},
  \bibinfo{author}{Zou, S.R.}, \bibinfo{author}{Cucchiara, R.},
  \bibinfo{author}{Tomasi, C.}, \bibinfo{year}{2016}.
\newblock \bibinfo{title}{Performance measures and a data set for multi-target,
  multi-camera tracking}.
\newblock \bibinfo{journal}{ECCV Workshops} .
\bibitem[{Romero et~al.(2015)Romero, Ballas, Kahou, Chassang, Gatta and
  Bengio}]{romero2015fitnets}
\bibinfo{author}{Romero, A.}, \bibinfo{author}{Ballas, N.},
  \bibinfo{author}{Kahou, E.S.}, \bibinfo{author}{Chassang, A.},
  \bibinfo{author}{Gatta, C.}, \bibinfo{author}{Bengio, Y.},
  \bibinfo{year}{2015}.
\newblock \bibinfo{title}{Fitnets: Hints for thin deep nets}.
\newblock \bibinfo{journal}{international conference on learning
  representations} .
\bibitem[{Russakovsky et~al.(2015)Russakovsky, Deng, Su, Krause, Satheesh, Ma,
  Huang, Karpathy, Khosla, Bernstein, Berg and Fei-Fei}]{RN991}
\bibinfo{author}{Russakovsky, O.}, \bibinfo{author}{Deng, J.},
  \bibinfo{author}{Su, H.}, \bibinfo{author}{Krause, J.},
  \bibinfo{author}{Satheesh, S.}, \bibinfo{author}{Ma, S.},
  \bibinfo{author}{Huang, Z.}, \bibinfo{author}{Karpathy, A.},
  \bibinfo{author}{Khosla, A.}, \bibinfo{author}{Bernstein, S.M.},
  \bibinfo{author}{Berg, C.A.}, \bibinfo{author}{Fei-Fei, L.},
  \bibinfo{year}{2015}.
\newblock \bibinfo{title}{Imagenet large scale visual recognition challenge}.
\newblock \bibinfo{journal}{International Journal of Computer Vision} .
\bibitem[{Sadeghian et~al.(2017)Sadeghian, Alahi and Savarese}]{RN455}
\bibinfo{author}{Sadeghian, A.}, \bibinfo{author}{Alahi, A.},
  \bibinfo{author}{Savarese, S.}, \bibinfo{year}{2017}.
\newblock \bibinfo{title}{Tracking the untrackable: Learning to track multiple
  cues with long-term dependencies}.
\newblock \bibinfo{journal}{arXiv preprint arXiv:1701.01909}
  \bibinfo{volume}{4}, \bibinfo{pages}{6}.
\bibitem[{Schaefer et~al.(2006)Schaefer, McPhail and Warren}]{RN988}
\bibinfo{author}{Schaefer, S.}, \bibinfo{author}{McPhail, T.},
  \bibinfo{author}{Warren, D.J.}, \bibinfo{year}{2006}.
\newblock \bibinfo{title}{Image deformation using moving least squares}.
\newblock \bibinfo{journal}{ACM Trans. Graph.} , \bibinfo{pages}{533--540}.
\bibitem[{Sheng et~al.(2018)Sheng, Chen, Zhang, Ke, Xiong and Yu}]{RN550}
\bibinfo{author}{Sheng, H.}, \bibinfo{author}{Chen, J.},
  \bibinfo{author}{Zhang, Y.}, \bibinfo{author}{Ke, W.},
  \bibinfo{author}{Xiong, Z.}, \bibinfo{author}{Yu, J.}, \bibinfo{year}{2018}.
\newblock \bibinfo{title}{Iterative multiple hypothesis tracking with
  tracklet-level association}.
\newblock \bibinfo{journal}{IEEE Transactions on Circuits and Systems for Video
  Technology} .
\bibitem[{Sun et~al.(2017)Sun, Zheng, Yang, Tian and Wang}]{RN985}
\bibinfo{author}{Sun, Y.}, \bibinfo{author}{Zheng, L.}, \bibinfo{author}{Yang,
  Y.}, \bibinfo{author}{Tian, Q.}, \bibinfo{author}{Wang, S.},
  \bibinfo{year}{2017}.
\newblock \bibinfo{title}{Beyond part models: Person retrieval with refined
  part pooling}.
\newblock \bibinfo{journal}{arXiv: Computer Vision and Pattern Recognition} .
\bibitem[{Tang et~al.(2017)Tang, Andriluka, Andres and
  Schiele}]{tang2017multiple}
\bibinfo{author}{Tang, S.}, \bibinfo{author}{Andriluka, M.},
  \bibinfo{author}{Andres, B.}, \bibinfo{author}{Schiele, B.},
  \bibinfo{year}{2017}.
\newblock \bibinfo{title}{Multiple people tracking by lifted multicut and
  person re-identification}.
\newblock \bibinfo{journal}{CVPR} , \bibinfo{pages}{3701--3710}.
\bibitem[{Urbann et~al.(2021)Urbann, Bredtmann, Otten, Richter, Bauer and
  Zibriczky}]{urbann2021online}
\bibinfo{author}{Urbann, O.}, \bibinfo{author}{Bredtmann, O.},
  \bibinfo{author}{Otten, M.}, \bibinfo{author}{Richter, J.P.},
  \bibinfo{author}{Bauer, T.}, \bibinfo{author}{Zibriczky, D.},
  \bibinfo{year}{2021}.
\newblock \bibinfo{title}{Online and real-time tracking in a surveillance
  scenario}.
\newblock \bibinfo{journal}{arXiv preprint arXiv:2106.01153} .
\bibitem[{Wang et~al.(2020)Wang, Wang, Lv, Hu and Li}]{RN1004}
\bibinfo{author}{Wang, H.}, \bibinfo{author}{Wang, S.}, \bibinfo{author}{Lv,
  J.}, \bibinfo{author}{Hu, C.}, \bibinfo{author}{Li, Z.},
  \bibinfo{year}{2020}.
\newblock \bibinfo{title}{Non-local attention association scheme for online
  multi-object tracking}.
\newblock \bibinfo{journal}{Image and Vision Computing} ,
  \bibinfo{pages}{103983}.
\bibitem[{Wang et~al.(2018)Wang, Girshick, Gupta and He}]{RN580}
\bibinfo{author}{Wang, X.}, \bibinfo{author}{Girshick, B.R.},
  \bibinfo{author}{Gupta, A.}, \bibinfo{author}{He, K.}, \bibinfo{year}{2018}.
\newblock \bibinfo{title}{Non-local neural networks}.
\newblock \bibinfo{journal}{CVPR} .
\bibitem[{Wen et~al.(2020)Wen, Du, Cai, Lei, Chang, Qi, Lim, Yang and
  Lyu}]{wen2020ua-detrac}
\bibinfo{author}{Wen, L.}, \bibinfo{author}{Du, D.}, \bibinfo{author}{Cai, Z.},
  \bibinfo{author}{Lei, Z.}, \bibinfo{author}{Chang, M.C.},
  \bibinfo{author}{Qi, H.}, \bibinfo{author}{Lim, J.}, \bibinfo{author}{Yang,
  M.H.}, \bibinfo{author}{Lyu, S.}, \bibinfo{year}{2020}.
\newblock \bibinfo{title}{Ua-detrac: A new benchmark and protocol for
  multi-object detection and tracking}.
\newblock \bibinfo{journal}{Computer Vision and Image Understanding} .
\bibitem[{Xu et~al.(2019)Xu, Cao, Zhang and Hu}]{xu2019spatial-temporal}
\bibinfo{author}{Xu, J.}, \bibinfo{author}{Cao, Y.}, \bibinfo{author}{Zhang,
  Z.}, \bibinfo{author}{Hu, H.}, \bibinfo{year}{2019}.
\newblock \bibinfo{title}{Spatial-temporal relation networks for multi-object
  tracking}.
\newblock \bibinfo{journal}{International Conference on Computer Vision} ,
  \bibinfo{pages}{3988--3998}.
\bibitem[{Yang et~al.(2016)Yang, Choi and Lin}]{yang2016exploit}
\bibinfo{author}{Yang, F.}, \bibinfo{author}{Choi, W.}, \bibinfo{author}{Lin,
  Y.}, \bibinfo{year}{2016}.
\newblock \bibinfo{title}{Exploit all the layers: Fast and accurate cnn object
  detector with scale dependent pooling and cascaded rejection classifiers}.
\newblock \bibinfo{journal}{CVPR} , \bibinfo{pages}{2129--2137}.
\bibitem[{You et~al.(2016)You, Wu, Li and Zheng}]{RN986}
\bibinfo{author}{You, J.}, \bibinfo{author}{Wu, A.}, \bibinfo{author}{Li, X.},
  \bibinfo{author}{Zheng, W.S.}, \bibinfo{year}{2016}.
\newblock \bibinfo{title}{Top-push video-based person re-identification}.
\newblock \bibinfo{journal}{CVPR} , \bibinfo{pages}{1345--1353}.
\bibitem[{Zhang et~al.(2018a)Zhang, Wu, Cheng, Zhang, Dong and Cai}]{RN969}
\bibinfo{author}{Zhang, D.}, \bibinfo{author}{Wu, W.}, \bibinfo{author}{Cheng,
  H.}, \bibinfo{author}{Zhang, R.}, \bibinfo{author}{Dong, Z.},
  \bibinfo{author}{Cai, Z.}, \bibinfo{year}{2018}a.
\newblock \bibinfo{title}{Image-to-video person re-identification with
  temporally memorized similarity learning}.
\newblock \bibinfo{journal}{IEEE Trans. Circuits Syst. Video Techn.} ,
  \bibinfo{pages}{2622--2632}.
\bibitem[{Zhang et~al.(2017)Zhang, Luo, Fan, Xiang, Sun, Xiao, Jiang, Zhang and
  Sun}]{RN984}
\bibinfo{author}{Zhang, X.}, \bibinfo{author}{Luo, H.}, \bibinfo{author}{Fan,
  X.}, \bibinfo{author}{Xiang, W.}, \bibinfo{author}{Sun, Y.},
  \bibinfo{author}{Xiao, Q.}, \bibinfo{author}{Jiang, W.},
  \bibinfo{author}{Zhang, C.}, \bibinfo{author}{Sun, J.}, \bibinfo{year}{2017}.
\newblock \bibinfo{title}{Alignedreid: Surpassing human-level performance in
  person re-identification}.
\newblock \bibinfo{journal}{Computer Vision and Pattern Recognition} .
\bibitem[{Zhang et~al.(2018b)Zhang, Xiang, Hospedales and Lu}]{RN983}
\bibinfo{author}{Zhang, Y.}, \bibinfo{author}{Xiang, T.},
  \bibinfo{author}{Hospedales, M.T.}, \bibinfo{author}{Lu, H.},
  \bibinfo{year}{2018}b.
\newblock \bibinfo{title}{Deep mutual learning}.
\newblock \bibinfo{journal}{CVPR} .
\bibitem[{Zhu et~al.(2017)Zhu, Jing, Wu, Wang, Zuo and Zheng}]{RN970}
\bibinfo{author}{Zhu, X.}, \bibinfo{author}{Jing, X.Y.}, \bibinfo{author}{Wu,
  F.}, \bibinfo{author}{Wang, Y.}, \bibinfo{author}{Zuo, W.},
  \bibinfo{author}{Zheng, W.S.}, \bibinfo{year}{2017}.
\newblock \bibinfo{title}{Learning heterogeneous dictionary pair with feature
  projection matrix for pedestrian video retrieval via single query image}.
\newblock \bibinfo{journal}{AAAI} , \bibinfo{pages}{4341--4348}.

\end{thebibliography}

%

\end{document}